\pdfoutput=1

\documentclass[11pt]{article}

\usepackage[preprint]{acl}

\usepackage{times}
\usepackage{latexsym}

\usepackage[T1]{fontenc}

\usepackage[utf8]{inputenc}

\usepackage{microtype}

\usepackage{inconsolata}

\usepackage[most]{tcolorbox}
\tcbset{
  promptbox/.style={
    colback=gray!10,    
    colframe=gray!50,   
    boxrule=0.5pt,
    arc=2pt,
    outer arc=2pt,
    boxsep=5pt,
    left=5pt,
    right=5pt,
    top=5pt,
    bottom=5pt,
    enhanced
  }
}

\usepackage{amsmath}
\usepackage{graphicx}
\usepackage{tikz}
\usepackage{pgfplots}
\usepackage{float}
\usepackage{colortbl}
\usetikzlibrary{shapes,arrows,positioning}
\pgfplotsset{compat=1.18}

\definecolor{badcolor}{RGB}{255,179,179} 
\definecolor{mehcolor}{RGB}{255,255,179} 
\definecolor{goodcolor}{RGB}{179,255,179} 
\definecolor{excellentcolor}{RGB}{102,204,102} 

\newcommand{\scite}[1]{{\scriptsize\cite{#1}}}

\newcommand{\scorecolor}[1]{%
  \ifdim #1pt<25pt
    \cellcolor{badcolor}#1
  \else\ifdim #1pt<50pt
    \cellcolor{mehcolor}#1
  \else\ifdim #1pt<75pt
    \cellcolor{goodcolor}#1
  \else
    \cellcolor{excellentcolor}#1
  \fi\fi\fi
}

\title{EasyMath: A 0-shot Math Benchmark for SLMs}

\author{
  Drishya Karki \\
  Assistantslab \\
  {\small\texttt{karkidrishya1@gmail.com}} \\[1ex]
  \And
  Michiel Kamphuis \\
  Assistantslab \\
  {\small\texttt{michielodevelopment@gmail.com}} \\[1ex]
  \And
  Angelecia Frey \\
  Assistantslab \\
  {\small\texttt{angeleciafrey@gmail.com}} \\[1ex]
}

\begin{document}
\maketitle
\begin{abstract}
EasyMath is a compact benchmark for practical math reasoning in small language models. It covers thirteen categories, from basic arithmetic and order of operations to word problems, algebraic expressions, edge cases, and omits specialist topics. We tested 23 models (14M to 4B parameters) using exact, numerical, and symbolic checks on free-form answers in a zero-shot setting. Accuracy rises with size and training, chain-of-thought adds modest gains, and consistency improves at scale.
\end{abstract}

\section{Introduction}
As small language models power mobile apps and chat assistants, evaluating their everyday math skills is crucial. Existing benchmarks either focus on highly complex academic problems or use multiple-choice formats that miss real problem-solving. EasyMath covers thirteen practical categories, from basic arithmetic and order of operations to word problems and edge cases, while skipping overly specialized topics. We score free-form answers with exact, numerical, and symbolic checks across models of different sizes and inference methods, revealing clear size and training effects, modest gains from chain-of-thought prompting, and consistency patterns to guide future work.

\section{Related Works}
The progress in both small language models (SLMs) and mathematical benchmarks has accelerated significantly in recent years, reflecting a broader trend towards specialized capabilities in natural language processing. In this section, we review the evolution of SLMs and the development of fitting mathematical benchmarks, providing the necessary background for understanding the contributions of our work.

\subsection{Progress in Small Language Models}
SLMs have traditionally been overshadowed by their larger counterparts in terms of capacity and performance. However, recent advances have demonstrated that, with proper architectural design and training strategies, smaller models can exhibit competitive performance on a range of tasks while retaining advantages such as reduced computational requirements and increased accessibility. For instance, the SmolLM2 models \cite{allal2025smollm2smolgoesbig} have shown that, even with fewer parameters, well-tuned SLMs can capture complex patterns in language data. This has been further reinforced by research exploring prompt engineering and inference strategies such as reasoning models that have shown to result in more accurate models \cite{zhang2024chain}.

We have also seen significant innovations in the design of efficient training regimes, including techniques such as knowledge distillation \cite{hinton2015distillingknowledgeneuralnetwork} and fine-tuning on domain-specific datasets \cite{mitra2024orcamathunlockingpotentialslms}. These approaches have enabled SLMs to achieve a level of performance in general language understanding and even in more specialized areas such as mathematical reasoning that is surprisingly strong given their limited scale, though still well behind the capabilities of larger models.

\subsection{Advances in Mathematical Benchmarks}
Parallel to the development of language models, the creation of mathematical benchmarks has played a pivotal role in assessing and pushing the boundaries of model reasoning capabilities. Early benchmarks focused on elementary arithmetic and algebraic tasks, providing a baseline for evaluating numerical reasoning. Datasets such as GSM8K \cite{cobbe2021trainingverifierssolvemath} and MathQA \cite{amini-etal-2019-mathqa} introduced more challenging problems that require the ability to interpret natural language in mathematical contexts and in some cases, perform multi-step reasoning.

More recent benchmarks have shifted the focus toward evaluating models across a wider spectrum of mathematical topics, ranging from basic arithmetic to more advanced subjects such as geometry, trigonometry, and complex algebraic expressions. These benchmarks have been instrumental in highlighting both the strengths and weaknesses of current models. Notably, research has shown that while large language models often benefit from chain-of-thought prompting \cite{wei2023chainofthoughtpromptingelicitsreasoning, kojima2023largelanguagemodelszeroshot}, SLMs sometimes struggle with such reasoning methods, revealing a critical gap in our understanding of how inference strategies interact with model size \cite{li2025smallmodelsstrugglelearn,kim2024smalllanguagemodelsequation,godey2024smalllanguagemodelsunderperform}.

Furthermore, complementary datasets that focus on edge-cases and word problems have been developed to test the robustness and generalizability of model performance. These datasets require not only mathematical proficiency but also strong language comprehension \cite{mirzadeh2024gsmsymbolicunderstandinglimitationsmathematical}. This dual focus has encouraged the development of evaluation methodologies, that provide a more holistic view of a model's capabilities.

\subsection{Bridging the Gap: From Benchmarks to Real-World Applications}
A common thread in the literature is the need to bridge the gap between benchmark performance and real-world applications. Although many benchmarks provide a controlled environment to test mathematical reasoning, they sometimes fail to capture the full complexity of everyday problem solving. This has led to a growing interest in designing benchmarks that are both comprehensive and practical. The work on EasyMath aims to address these challenges by creating a dataset that not only covers a wide range of mathematical topics but also reflects the types of problems encountered in daily use to better understand the strengths and weaknesses of a model.

By integrating insights from both the advancements in SLMs and the evolution of mathematical benchmarks, our work contributes to a more nuanced understanding of model performance. We build on the extensive literature in these fields, leveraging recent techniques in prompt engineering, symbolic reasoning, and evaluation strategies to propose a benchmark tailored to the strengths and limitations of SLMs.

In summary, the intersection of SLM research and mathematical benchmarking presents a fertile ground for exploring the capabilities of modern language models. As SLMs become increasingly relevant for resource-constrained applications, our focus on developing and evaluating benchmarks like EasyMath is essential for ensuring that these models are robust, interpretable, and effective in real-world scenarios.

\section{Reflection on Current Math Benchmarks}
The current landscape of mathematical benchmarks has driven significant progress in assessing the reasoning capabilities of language models. However, a closer examination reveals several limitations that necessitate the development of alternative evaluation methodologies, such as the EasyMath benchmark. In this section, we critically assess the shortcomings of existing benchmarks for SLMs, highlighting the gap between controlled evaluation settings and the demands of real-world applications.

\subsection{Challenges with GSM8K}
GSM8K \cite{cobbe2021trainingverifierssolvemath} has emerged as a popular benchmark for mathematical problem solving, yet it is often characterized as being excessively challenging, especially for SLMs. Empirical observations show that SLMs frequently register a near-zero accuracy in a zero-shot setting on GSM8K. Consequently, models often resort to multi-shot evaluations (e.g., 8-shot prompting) to achieve what is seen as "acceptable" scores, a practice that may not accurately reflect their true, standalone reasoning capabilities in practical settings.

\subsection{Limitations of Multiple-Choice Frameworks in MathQA}
MathQA \cite{amini-etal-2019-mathqa} adopts a multiple-choice format coupled with aligned operation programs to evaluate mathematical proficiency. While this design can effectively isolate specific computational abilities, it introduces an artificial level of precision. In realistic problem-solving scenarios, users are rarely provided with a set of predetermined options; instead, they must interpret and solve open-ended problems. This discrepancy can lead to an overestimation of a model’s practical competence in handling free-form mathematical reasoning and complex problem descriptions.

\subsection{Gaps in Hendrycks-MATH and Math-500}
Hendrycks-MATH \cite{hendrycks2021measuringmathematicalproblemsolving} represents a comprehensive attempt to cover a broad spectrum of mathematical topics—from prealgebra and algebra to geometry and number theory. However, it was deliberately designed to be challenging, and notably, it omits critical areas such as large-number arithmetic. Similarly, Math-500 focuses on foundational subjects like precalculus and intermediate algebra but overlooks certain edge cases and complex real-world problem structures. Their specific focus and difficulty limits the utility of such benchmarks for evaluating SLMs intended for everyday use.

\section{EasyMath benchmark}
The EasyMath benchmark addresses the critical gap in evaluating SLMs for practical mathematical reasoning by focusing on real-world applicability and incremental reasoning capabilities. Unlike existing benchmarks which prioritize complexity or niche problem types, EasyMath is designed to assess foundational and intermediate mathematical skills that are essential for everyday use. By emphasizing categories such as word problems, percentages, order of operations, and multi-step reasoning, while avoiding overly complex topics like linear algebra, the benchmark provides a nuanced evaluation of SLMs’ ability to handle the types of tasks users encounter daily. This structured approach, combined with a flexible evaluation strategy that accounts for both exact and symbolic equivalence, ensures that EasyMath is a robust benchmark for measuring the strengths and limitations of SLMs.

\subsection{Methodology}
With the goal of creating a categorized dataset that contains relatively easy math question, we decided to include 13 different categories. To avoid more complex math, we did not include topics such as linear algebra and started out with extremely basic arithmetic and worked our way up to multi-step-reasoning. Whereas basic arithmetic contains questions such as "What is 5 plus 3?", multi-step-reasoning includes some harder questions where the models must take a number of steps to find the answer such as "A shirt costs \$40. It is first discounted by 20\%, then by an additional 10\%. What is the final price?". This way, we can get a better understanding of the mathematical reasoning and capabilities of a model.

\subsubsection{Categories}
We want to specifically focus on a number of categories within mathematics that will be useful to a SLM in everyday use, such as basic addition, subtraction, fractions, decimals, and percentages. We also wanted to test if a model is able to perform calculations that require knowledge about order of operations. Additionally, we can test how well a model can use exponents, roots, geometry, and trigonometry. These topics are more advanced than basic arithmetic and require deeper mathematical understanding. Finally, we introduce large-numbers, word-problems, complex sentences, algebraic expressions, edge-cases, and multi-step-reasoning. 

\subsubsection{Dataset Development and Curation}
Having established what categories we want to include in EasyMath, we proceeded to create the dataset by writing both the questions and the corresponding solutions manually. To once again focus on real-world performance, we chose to avoid multiple-choice questions such as the MathQA benchmark \cite{amini-etal-2019-mathqa}. Following the manual creation, we ensured no questions overlapped with various existing math benchmarks and training datasets. Furthermore, for each question, we added the calculation that could be performed using SymPy \cite{10.7717/peerj-cs.103} to arrive at the correct answer. Some mathematical datasets include multiple correct answers, as solutions can often be expressed in multiple ways \cite{yue2024harp}. Instead of taking this approach, we determine if an answer is equivalent to the answer we provided during the evaluation itself.

\subsection{Evaluation}
The EasyMath evaluation pipeline (flowchart in Appendix B, Figure \ref{fig:evaluation_flowchart}) comprises the following stages:
\begin{enumerate}
    \item \textbf{Response Processing}: The model's response is processed to extract all potential mathematical expressions using regular expression patterns.
    
    \item \textbf{Normalization}: Extracted expressions undergo normalization to standardize notation (converting symbols like $\sqrt{x}$ to sqrt(x), handling superscripts, etc.) and ensure consistent formatting.
    
    \item \textbf{Category-Based Evaluation}: The evaluation strategy varies based on the problem category:
    \begin{itemize}
        \item For categories requiring exact answers (basic arithmetic, order of operations, large numbers, and exponents/roots), a strict matching approach is used.
        \item For other categories (algebra, geometry, word problems, etc.), a more flexible equivalence-based approach is employed.
    \end{itemize}
    
    \item \textbf{Matching Methods}: Several methods are used to determine if an extracted expression matches the expected result:
    \begin{itemize}
        \item \textbf{Direct String Matching}: Exact string comparison after normalization.
        \item \textbf{Numerical Evaluation}: Converting expressions to numerical values for comparison.
        \item \textbf{Symbolic Equivalence}: Using SymPy \cite{10.7717/peerj-cs.103} to determine if expressions are mathematically equivalent.
        \item \textbf{Difference Simplification}: Checking if the symbolic difference between expressions simplifies to zero.
    \end{itemize}
    
    \item \textbf{Result Determination}: If any matching method succeeds, the response is marked as correct; otherwise, it's marked as incorrect.
\end{enumerate}

This multi-layered approach ensures robust evaluation across diverse mathematical problem types and accounts for variations in expression format and notation. The only exception to this standard process is for the edge-cases category, where the prompt includes the instruction to explicitly respond with "undefined" if the question cannot be solved. For this category, we expect "undefined" to be in the model response for it to be evaluated as correct.

\subsection{Confidence interval of intended answer}

In our evaluation of the EasyMath benchmark, we have determined that the phenomenon of a model accidentally mentioning the correct answer, even when the overall reasoning is incorrect, must be taken into account or at least analyzed. Our analysis is based on the following definitions:

\begin{itemize}
    \item Let \( Q \) be the set of values explicitly present in the problem statement.
    \item Let \( A_c \) be the correct answer for a given problem.
    \item Let \( I \) be the set of mathematically relevant intermediate values in typical solution paths.
\end{itemize}

\subsubsection*{Probability Model}

For a model that arrives at an incorrect conclusion, the probability of the correct answer appearing accidentally is given by:
\begin{align}
P(A_c \in R \mid A_c \notin C) =\ & P(A_c \in Q) \nonumber \\
&\hspace{-8em} + P(A_c \in I) \cdot \left(1 - P(A_c \in Q)\right)
\end{align}

The probabilities \( P(A_c \in Q) \) and \( P(A_c \in I) \) are estimated based on heuristic analysis of the problem type. For example, word problems might have a higher chance of including contextual values directly (e.g., \( P(A_c \in Q) \approx 0.05 \)) compared to computation problems where \( P(A_c \in Q) = 0 \).

\subsection*{Empirical Findings}

Using the EasyMath benchmark, we computed the accidental mention probability for each problem. Averaging across all problems, the \textbf{average accidental mention probability}\footnote{
We parsed each question and its associated calculation to extract numeric tokens, flagged any that matched the final answer or plausible intermediates, applied category‐specific priors \(P(A_c\in Q)\) and \(P(A_c\in I)\), computed each problem’s accidental mention probability and averaged across all problems.
} was found to be:
\[
\bar{P}(A_c \in R \mid A_c \notin C) = 0.0419
\]

\subsection*{Adjusted Accuracy for Llama-3.2-1B}

Given that Llama-3.2-1B has a reported accuracy of \( a = 74.92\% \) in table \ref{tab:benchmark_results}, we can compute an adjusted accuracy to account for accidental mentions using:
\begin{equation}
    a_{\text{adjusted}} = a - (1 - a) \cdot \bar{P}(A_c \in R \mid A_c \notin C)
\end{equation}

Substituting the values:
\begin{align*}
    a_{\text{adjusted}} &= 0.7492 - (1 - 0.7492) \cdot 0.0419 \\
    &= 0.7492 - 0.2508 \cdot 0.0419 \\
    &= 0.7492 - 0.01050852 \\
    &\approx 0.7387 \quad (\text{or } 73.87\%)
\end{align*}

\subsection*{Interpretation and Confidence Margin}

The analysis shows that accidental mentions occur in approximately 4\% of cases, leading to a modest reduction in apparent accuracy (from 74.92\% to about 73.87\%). However, given the relatively low probability and the inherently probabilistic nature of these estimates, it is fair to interpret this difference as a confidence margin rather than a strict adjustment to the reported scores. In practical terms, while the adjusted accuracy offers insight into the evaluation robustness, the original score remains a valid performance measure with an understanding that there is an inherent margin of error on the order of approximately 1--2 percentage points.

\section{Results \& Learning to overcome EasyMath}
The EasyMath benchmark provides a comprehensive evaluation of large and small language models' ability to perform practical mathematical reasoning. We present models tested on EasyMath ranging from 14 million parameters up until 4 billion parameters. The results highlight the varying levels of success models achieve across different categories. In addition, we analyzed various factors in these results such as differing inference strategies, model consistency, and ultimately compare EasyMath against existing benchmarks to see which benchmark can provide the best snapshot of models' mathematical accuracy and understanding.

\subsection{Results}
We benchmarked a total of 23 models, from standard instruct-tuned models to specialized math-tuned models. We also include some small reasoning models, as these models have recently been shown to improve skills such as math. Our expectation is that models under 1B parameters may struggle with most subjects, whereas models above 1B parameters start to get more accurate. It is important to note that our EasyMath questions include topics such as basic arithmetic, which is taught in elementary school. As a result, we expect models to answer these simple questions correctly, leading to a non-zero average accuracy.

All models have been tested with appropriate prompting methods, either based on model-specific information or our own testing. For example, the Pythia \cite{biderman2023pythiasuiteanalyzinglarge} model series are purely text-completion based and use a simple Q-A format. Furthermore, all tests were fully 0-shot, which does introduce some randomization that should be kept in mind when interpreting these results.

\begin{table}[H]
    \centering
    \begin{tabular}{|c|c|}
        \hline
        \textbf{Model} & \textbf{Avg. score} \\
        \hline
        pythia-14m \scite{biderman2023pythiasuiteanalyzinglarge} & \scorecolor{1.85} \\
        \hline
        gpt-neo-125m \scite{gpt-neo} & \scorecolor{8.15} \\
        \hline
        SmolLM2-135M \scite{allal2025smollm2smolgoesbig} & \scorecolor{28.77} \\
        \hline
        pythia-160m \scite{biderman2023pythiasuiteanalyzinglarge} & \scorecolor{6.77} \\
        \hline
        SmolLM2-360M \scite{allal2025smollm2smolgoesbig} & \scorecolor{48.77} \\
        \hline
        pythia-410m \scite{biderman2023pythiasuiteanalyzinglarge} & \scorecolor{9.54} \\
        \hline
        Qwen2.5-0.5B \scite{qwen2025qwen25technicalreport} & \scorecolor{88.31} \\
        \hline
        Falcon3-1B \scite{Falcon3} & \scorecolor{77.69} \\
        \hline
        Llama-3.2-1B \scite{grattafiori2024llama3herdmodels,meta_llama} & \scorecolor{74.92} \\
        \hline
        gpt-neo-1.3B \scite{gpt-neo} & \scorecolor{13.85} \\
        \hline
        pythia-1.4b \scite{biderman2023pythiasuiteanalyzinglarge} & \scorecolor{16.77} \\
        \hline
        DeepScaleR-1.5B \scite{deepscaler2025} & \scorecolor{90.00} \\
        \hline
        Qwen2.5-1.5B \scite{qwen2025qwen25technicalreport} & \scorecolor{93.08} \\
        \hline
        AceMath-1.5B \scite{liu2025acemathadvancingfrontiermath} & \scorecolor{95.69} \\
        \hline
        SmolLM2-1.7B \scite{allal2025smollm2smolgoesbig} & \scorecolor{80.62} \\
        \hline
        Gemma-3-4B \scite{Gemma3Report2025} & \scorecolor{91.38} \\
        \hline
    \end{tabular}
    \caption{Abbreviated benchmark results (models sorted by parameter count). For the complete set of models and all benchmark categories, refer to appendix C, table 2.}
    \label{tab:benchmark_results_shortened}
\end{table}

In the current model landscape, we can observe a moderate correlation that confirms larger models tend to score higher, but it also highlights that size alone cannot fully predict accuracy. Small models like SmolLM2 (135M) and Qwen2.5 (0.5B) fall well below the trend line, whereas Falcon 3 (1B) and Llama 3.2 (1B) adhere almost exactly to it. Mid-sized architectures (e.g., Gemma-3 1B, DeepScaleR 1.5B, AceMath 1.5B) cluster more tightly, suggesting that at these scales, design, and training regimen start to dominate over pure parameter count.

The patterns we can find in appendix C, table \ref{tab:benchmark_results}, is that algebraic expressions (AE) and large‑number (LN) tasks remain a hurdle for smaller and mid‑sized models. Sub‑200M‑parameter models uniformly score below 25\% on AE and essentially 0\% on LN, reflecting both the probabilistic weaknesses of LLMs in handling multi‑digit arithmetic and the fragility revealed by GSM‑Symbolic when even trivial value perturbations collapse performance \cite{mirzadeh2024gsmsymbolicunderstandinglimitationsmathematical}. This limitation is exacerbated by the small models’ inability to internalize long chain‑of‑thoughts: direct distillation of complex reasoning often fails to benefit them, and they instead perform better with shorter, simpler chains \cite{li2025smallmodelsstrugglelearn}. Converting problems into an equation‑only format can reduce natural‑language ambiguity but does not fully close the gap \cite{kim2024smalllanguagemodelsequation}, and inherent capacity saturation further caps their progress \cite{godey2024smalllanguagemodelsunderperform}.

Mid‑tier architectures (350M–1B parameters) make measurable gains. SmolLM2‑360M reaches 36\% AE and 12\% LN, while Qwen2.5‑0.5 B jumps to 78\% AE and 48\% LN, but still fall short of reliable math proficiency. Only at or above the 1.5B‑parameter scale do models begin to break through: instruction‑tuned and reasoning‑distilled variants (e.g., DS‑R1‑Distill‑Qwen‑1.5B, Qwen2.5‑Math‑1.5B) achieve near‑perfect AE (98–100\%) and substantial LN accuracy (40–72\%). AceMath‑1.5 leads with 100\% AE and 76\% LN, and Falcon3‑1B‑Instruct stands out among 1B‑parameter models with 82\% AE while Gemma-3 1B stands out for its 68\% score on LN, underscoring that targeted reasoning‑focused training is key to unlocking robust mathematical capabilities \cite{Falcon3,Gemma3Report2025}.

\subsection{Consistency in responses}
To assess the consistency of LLMs on EasyMath, we selected three models with varying parameter sizes and ran our benchmark 10 times on each model. The models we tested were SmolLm2-135M \cite{allal2025smollm2smolgoesbig}, Llama-1B \cite{meta_llama}, and Gemma3-4B \cite{Gemma3Report2025}.

\subsubsection{Trend Analysis}
We find notable differences in run-to-run variability between models. The standard deviation in overall accuracy across the 10 runs shows a pattern related to model size, suggesting that larger models exhibit more consistent performance across multiple runs.

\begin{itemize}
\item Gemma-3-4B (4B parameters): ±0.30\% (range: 90.54\%-91.35\%)
\item Llama-3.2-1B (1B parameters): ±1.11\% (range: 70.81\%-74.86\%)
\item SmolLM2-135M (135M parameters): ±1.71\% (range: 20.81\%-26.76\%)
\end{itemize}

These findings align with \citet{lyu2024calibratinglargelanguagemodels}, who demonstrated that scaling up model size enhances calibration and consistency across various reasoning tasks. Their work suggests that extensive sampling may not be necessary for reliable evaluations.

\subsubsection{Category-Specific Variability}
When examining consistency across mathematical categories, we found that different types of mathematical problems exhibit varying degrees of run-to-run consistency. We find that smaller models generally exhibit higher variability across most mathematical categories (see Appendix D, Figure \ref{fig:category_variability_heatmap} for a comprehensive heatmap illustrating these differences).

\subsubsection{Statistical Reliability and Confidence Intervals}
To establish the statistical validity of our benchmark assessments, we calculated 95\% confidence intervals (CIs) for each model's accuracy using the t-distribution:

\begin{itemize}
    \item Gemma-3-4B: 90.84\% [90.62\%, 91.05\%] (CI width: 0.47\% of mean)
    \item Llama-3.2-1B: 73.05\% [72.26\%, 73.85\%] (CI width: 2.17\% of mean)
    \item SmolLM2-135M: 24.08\% [22.86\%, 25.30\%] (CI width: 10.15\% of mean)
\end{itemize}

These confidence intervals indicate that accuracy measurements for the larger models are highly representative of their true capabilities, with relative CI widths well below 5\% of their means.
We also assessed whether a single evaluation run provides a reasonable estimate of model capabilities through bootstrap analysis with 10,000 resamples. The mean error from a single run (compared to the mean across all runs) was 0.23\% for Gemma-3-4B, 1.01\% for Llama-3.2-1B, and 5.32\% for SmolLM2-135M. Our analysis suggests that larger models can be reliably evaluated with fewer runs, while smaller models benefit more from multiple evaluations.

These results are consistent with \citet{wang2025assessingconsistencyreproducibilityoutputs}. They found that simple aggregation strategies across just 3-5 runs significantly improved consistency, and despite some run-to-run variations, downstream statistical inferences remained robust. Their work supports our finding that extensive runs may not be necessary to achieve reliable evaluations, particularly for larger models.

\subsection{Strategies to overcome EasyMath}
\begin{figure}[H]
\centering
\includegraphics[width=0.5\textwidth]{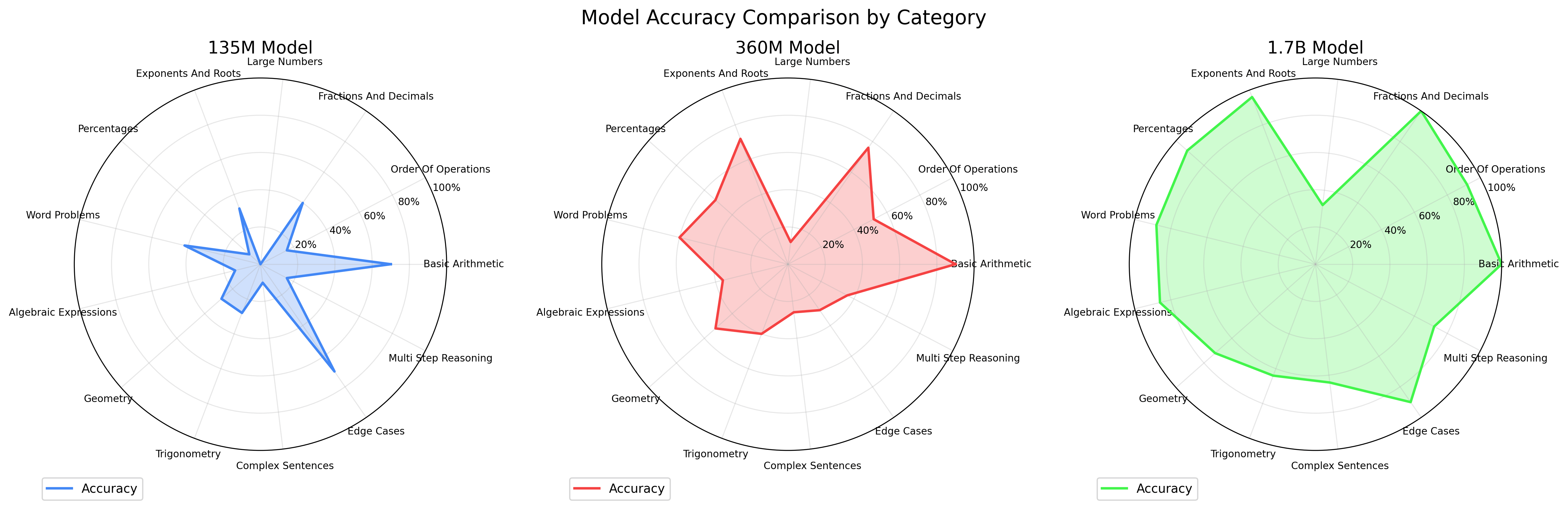}
\caption{SmolLM2 models accuracy}
\label{fig:smollm2_series_accuracy}
\end{figure}

EasyMath is ultimately a benchmark designed for smaller models to complete successfully, due to the simplicity of the questions and computations. Parameter size alone is a significant factor in improving performance on EasyMath, as illustrated by figure \ref{fig:accuracy_vs_param_log} and further detailed in the comparison of three different SmolLM2 models in figure \ref{fig:smollm2_series_accuracy}. That being said, considering that having better SLMs is valuable for a plethora of use-cases, we can analyze strategies to improve the accuracy of a given model without increasing the amount of parameters.

\subsubsection{The effect of training on existing datasets}
We examined the impact of continued finetuning on SmolLM2 models by training them on established mathematics datasets and evaluating their performance on EasyMath. Two separate models were trained for 8 epochs each on subsets of prominent mathematics datasets: 25,000 randomly sampled examples from Orca-word-problems \cite{mitra2024orcamathunlockingpotentialslms} and 25,000 randomly sampled examples from OpenR1-Math-220k \cite{openr1,OpenR1-Math-220k}. The former dataset is a dataset that specifically focuses on word problems, whereas the latter focuses on reasoning through various math problems. We observe a quickly-reached plateau that seems dependent on the parameter size of the model (appendix E, figure \ref{fig:accuracy_plot_finetunes}). This indicates that existing datasets can already quite significantly improve mathematical understanding and reasoning, especially with the orca-word-problems \cite{mitra2024orcamathunlockingpotentialslms} dataset we can observe an improvement of roughly 12\%. In general, the orca-word-problems dataset is slightly higher quality than the OpenR1-Math-220k dataset which could indicate higher-quality data makes a relatively big impact on any changes in model performance.

\subsubsection{Different inference strategies}
Recently we have seen a new popular approach to improve mathematical knowledge amongst other fields and benchmarks by just reasoning about a given prompt and working through problems step by step. This makes sense given the probabilistic nature of LLMs and SLMs. In our testing, we therefore included the reasoning models Deepseek-R1-Distill-Qwen-1.5B and DeepScaleR-1.5B. We would have expected that these reasoning models may have been better compared to the other tested models, but these models instead align with the average expectation for their size. Even when compared specifically to other 1.5B models, they are not actually better. This suggests that in SLMs, reasoning through the problem is not enough to solve mathematical problems with a higher than normal accuracy. However, this leaves some questions unanswered, specifically the question of why reasoning in SLMs does not significantly improve the accuracy of models, even though such behavior is expected in LLMs \cite{yu2025thinksmarterharderadaptive}.

It may be good to differentiate between chain-of-thought reasoning and the fundamental reasoning models. For example, the aforementioned reasoning models. Chain-of-thought reasoning refers to the practice of simply reasoning through problems \cite{wei2023chainofthoughtpromptingelicitsreasoning, kojima2023largelanguagemodelszeroshot}, a number of papers explicitly show improving accuracy on mathematical tasks \cite{wei2022chainofthought,cobbe2021trainingverifierssolvemath,hendrycks2021measuringmathematicalproblemsolving,lightman2023letsverifystepstep}. Most models tested use this approach to solve math problems to increase accuracy, which we see back in the results. Some good examples are the SmolLM2 model series \cite{allal2025smollm2smolgoesbig} and the Qwen2.5 model series \cite{qwen2025qwen25technicalreport,yang2024qwen25mathtechnicalreportmathematical} amongst others, compared to the Pythia \cite{biderman2023pythiasuiteanalyzinglarge} or gpt-neo \cite{gpt-neo} model series. The former models show a strong improvement on the latter models, and although the training data and architecture of the models definitely plays a role, the chain-of-thought reasoning is a critical factor in this improvement. We can verify this by 'prompt engineering' \cite{schulhoff2025promptreportsystematicsurvey,wei2022chainofthought} and highly encouraging the latter models (Pythia and gpt-neo series) to attempt to reason through problems. The expectation would be that the Pythia and gpt-neo series would improve their accuracy when being prompt engineered but they still underperform compared to other models, due to these other models being trained for CoT prompting. This expectation is in line with the observed behavior in appendix G, table \ref{tab:pythia-easymath-scores}, where almost all categories achieve a higher accuracy when using a CoT prompt. The average accuracy improves from 16.77 with standard question-answer prompting to 22.92 with CoT prompting, which is an improvement of 6.15\%.

Our findings are in line with existing literature and we clearly observe that reasoning through problems results in models being more accurate \cite{srivastava2025reasoningabilitysmalllanguage}. Whereas reasoning LLMs often outperform regular LLMs with CoT prompting, we do not see this behavior in EasyMath, which is also in line with existing literature. It seems that SLMs have a harder time correctly reasoning compared to LLMs, resulting in roughly the same accuracy as regular models with CoT prompting \cite{wei2022chainofthought,cobbe2021trainingverifierssolvemath,hendrycks2021measuringmathematicalproblemsolving,lightman2023letsverifystepstep,wei2023chainofthoughtpromptingelicitsreasoning, kojima2023largelanguagemodelszeroshot}. Interestingly enough, the same existing literature foresees a future where SLMs with strong reasoning capabilities can be developed through structured training or post-training compression that will outperform regular models with CoT prompting \cite{srivastava2025reasoningabilitysmalllanguage}, which could indicate that the current reasoning SLMs are simply not yet at the accuracy and quality that they could be in the future.

\subsection{Comparison to existing benchmarks}
To compare Easymath with other benchmarks, we tested six different language models ranging from 70 million to 3 billion parameters in a 0-shot setting. Across the table, GSM8K had the overall lowest performance, as it did not return any useful results across all models, which severely limits its usefulness for comparing SLMs. Both Hendrycks-Math and Math-500 began showing nonzero scores around the 0.5B parameter range. Besides having zero scores in the models below the 0.5B mark these other scores help provide us some insight into a proper comparison between models. 

We should keep in mind that when multiple models all score zero, that benchmark practically becomes almost useless for comparison. It no longer tells us how the models differ or how well they are preforming. This doesn't mean the benchmark is necessarily bad, just that it may not be well suited for the type of models being tested. Some benchmarks are designed with much larger models in mind that don't register zero scores.

MathQA showed nonzero scores across all models, but it is also a multiple choice benchmark by design, meaning models are likely just guessing. With four options per question, random guessing would land around 25\%, and the scores that we saw had an average deviation of just 4.29\% compared to that, which limits how much we can truly learn from it. On the other hand, compared to the other benchmarks (GSM8K, Hendrycks-Math, Math-500, and MathQA) EasyMath was the only benchmark to consistently give nonzero scores without relying on a multiple choice approach. This makes EasyMath more reliable for measuring and comparing SLMs in a meaningful way.

\section{Conclusion}
In this work, we introduced EasyMath, a lightweight yet comprehensive benchmark designed to evaluate the real-world mathematical reasoning capabilities of small language models. By focusing on everyday problem types—ranging from basic arithmetic and percentages to multi-step word problems and edge cases—we provide a targeted lens through which to assess a model’s practical utility. Our multi-layered evaluation pipeline, combining exact matching, numerical comparison, and symbolic equivalence checks, ensures both rigor and flexibility in grading open-ended responses.

Empirical results across 23 models (14M–4B parameters) reveal clear trends: larger models generally achieve higher accuracy, but architectural design and training regimen play an equally important role, especially in the 350 M–1 B range. We further demonstrate that chain-of-thought prompting boosts performance for small models—albeit not to the same extent observed in LLMs—and quantify the modest impact of accidental answer mentions on reported accuracies. Consistency analyses confirm that reliability improves with scale, underscoring the importance of multiple runs for smaller architectures.

Looking ahead, EasyMath lays the groundwork for several promising directions. First, integrating dynamic, user-generated problem sets could further bridge the gap between benchmarks and live usage. Second, exploring hybrid inference strategies—combining symbolic solvers with lightweight neural modules—may unlock stronger performance in resource-constrained settings. Finally, extending our accidental-mention framework to other reasoning domains could refine evaluation protocols across NLP tasks.

By aligning evaluation more closely with everyday demands and highlighting the nuanced interplay between size, training, and prompting, EasyMath offers both a practical tool and a roadmap for future research on expressive yet efficient language models.

\section{Limitations}
Our work focuses on SLMs and is therefore scoped accordingly. Larger models are likely to score near-perfect accuracy on EasyMath, making it less informative; other benchmarks (e.g., GSM8K \cite{cobbe2021trainingverifierssolvemath}) target more complex problems. Additionally, the dataset was manually constructed to ensure quality and relevance, which may introduce subtle linguistic biases despite contributions from multiple authors.

Our evaluation pipeline relies on symbolic matching using SymPy \cite{10.7717/peerj-cs.103}. Although we implement additional checks—direct numerical matching, string equivalence, and input normalization—some expressions may still evade recognition depending on formatting or notation. We have not observed this for the models evaluated here, but other models’ outputs might require explicit formatting instructions.

The benchmark is limited to English-language prompts and does not assess multilingual performance or reasoning over non-text modalities.

Finally, although we use prompt formats recommended or tested per model family (e.g., Q–A for Pythia \cite{biderman2023pythiasuiteanalyzinglarge}), performance may vary with different prompting strategies, especially on models not optimized for reasoning out of the box.

\bibliography{custom}

\clearpage

\appendix
\onecolumn

\section{Question distribution across categories}
\label{sec:appendix}
Since not all of these skills contribute equally to a model’s real-world performance, the distribution of questions per category is not uniform. More representation is given to categories that are particularly challenging or important for small models in a conversational setting. Algebraic expressions are for example, more represented compared to other categories due to the difficulty it seems to pose to SLMs and their stark difference from basic arithmetic up to simple word problems \cite{kim2024smalllanguagemodelsequation}.
\nopagebreak
\begin{figure}[H]
    \centering
    \begin{tikzpicture}
        \begin{axis}[
            ybar,
            width=0.9\textwidth,
            height=0.3\textwidth,
            xlabel={Categories},
            ylabel={Number of Questions},
            symbolic x coords={
                basic\_arithmetic,
                edge\_cases,
                order\_of\_operations,
                fractions\_and\_decimals,
                large\_numbers,
                exponents\_and\_roots,
                percentages,
                geometry,
                trigonometry,
                multi\_step\_reasoning,
                word\_problems,
                algebraic\_expressions,
                complex\_sentences
            },
            xtick=data,
            x tick label style={rotate=45, anchor=east},
            nodes near coords,
            enlarge x limits=0.05,
            ymin=0, ymax=60,
            grid=both,
            major grid style={gray!25},
            minor grid style={gray!15},
        ]
        \addplot coordinates {
            (basic\_arithmetic,10)
            (edge\_cases,10)
            (order\_of\_operations,25)
            (fractions\_and\_decimals,25)
            (large\_numbers,25)
            (exponents\_and\_roots,25)
            (percentages,25)
            (geometry,25)
            (trigonometry,25)
            (multi\_step\_reasoning,25)
            (word\_problems,50)
            (algebraic\_expressions,50)
            (complex\_sentences,50)
        };
        \end{axis}
    \end{tikzpicture}
    \caption{Distribution of Questions Across Categories}
    \label{fig:category_distribution}
\end{figure}
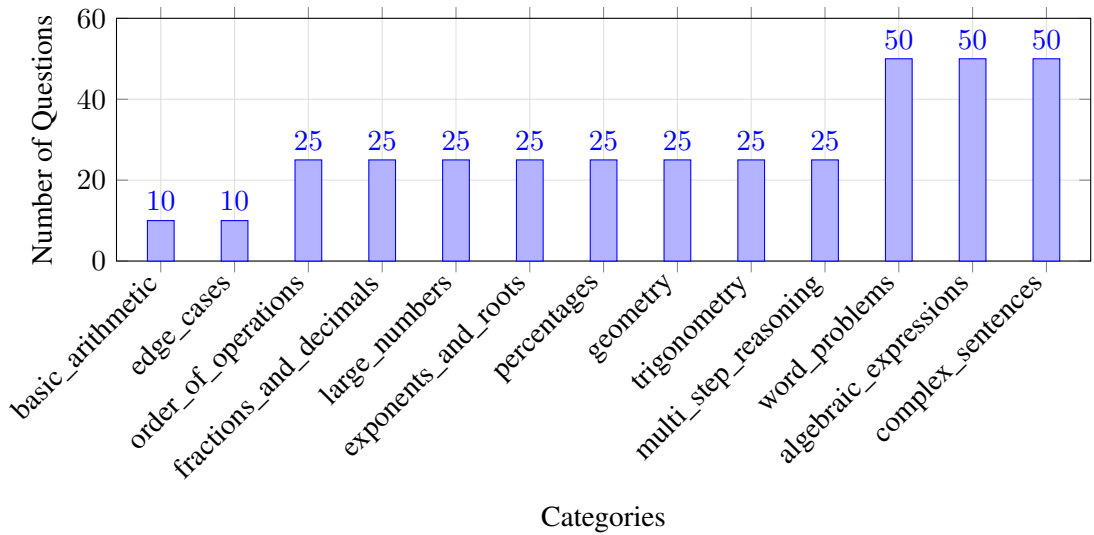

Below are some representative example questions from selected categories:

\begin{tcolorbox}[enhanced,
                  colback=gray!10,
                  colframe=gray!50,
                  fonttitle=\bfseries,
                  title=Example Questions by Category]

\textbf{Basic Arithmetic}: \\
Add 7 and 6 together.

\vspace{0.5em}
\textbf{Order of Operations}: \\
What is \((12 + 4\)) divided by \((5 - 3\))?

\vspace{0.5em}
\textbf{Fractions and Decimals}: \\
Divide \( \frac{3}{4} \) by \( \frac{1}{2} \).

\vspace{0.5em}
\textbf{Word Problems}: \\
How long does it take to drive 180 miles at 60 mph?

\vspace{0.5em}
\textbf{Algebraic Expressions}: \\
Simplify \( \frac{x^5}{x^2} \)

\vspace{0.5em}
\textbf{Geometry}: \\
Find the circumference of a circle with radius 7.

\vspace{0.5em}
\textbf{Complex Sentences}: \\
Triple the sum of 5 and 7, then subtract the square of 4.

\vspace{0.5em}
\textbf{Edge cases}: \\
Solve for x: 0x = 5. Respond with Undefined if you can't answer it, but try your best.

\vspace{0.5em}
\textbf{Multi-step-reasoning}: \\
A recipe requires 3 parts flour to 2 parts sugar. If you use 12 cups of flour, how much sugar is needed?

\end{tcolorbox}
\clearpage

\section{Evaluation Process}
\nopagebreak
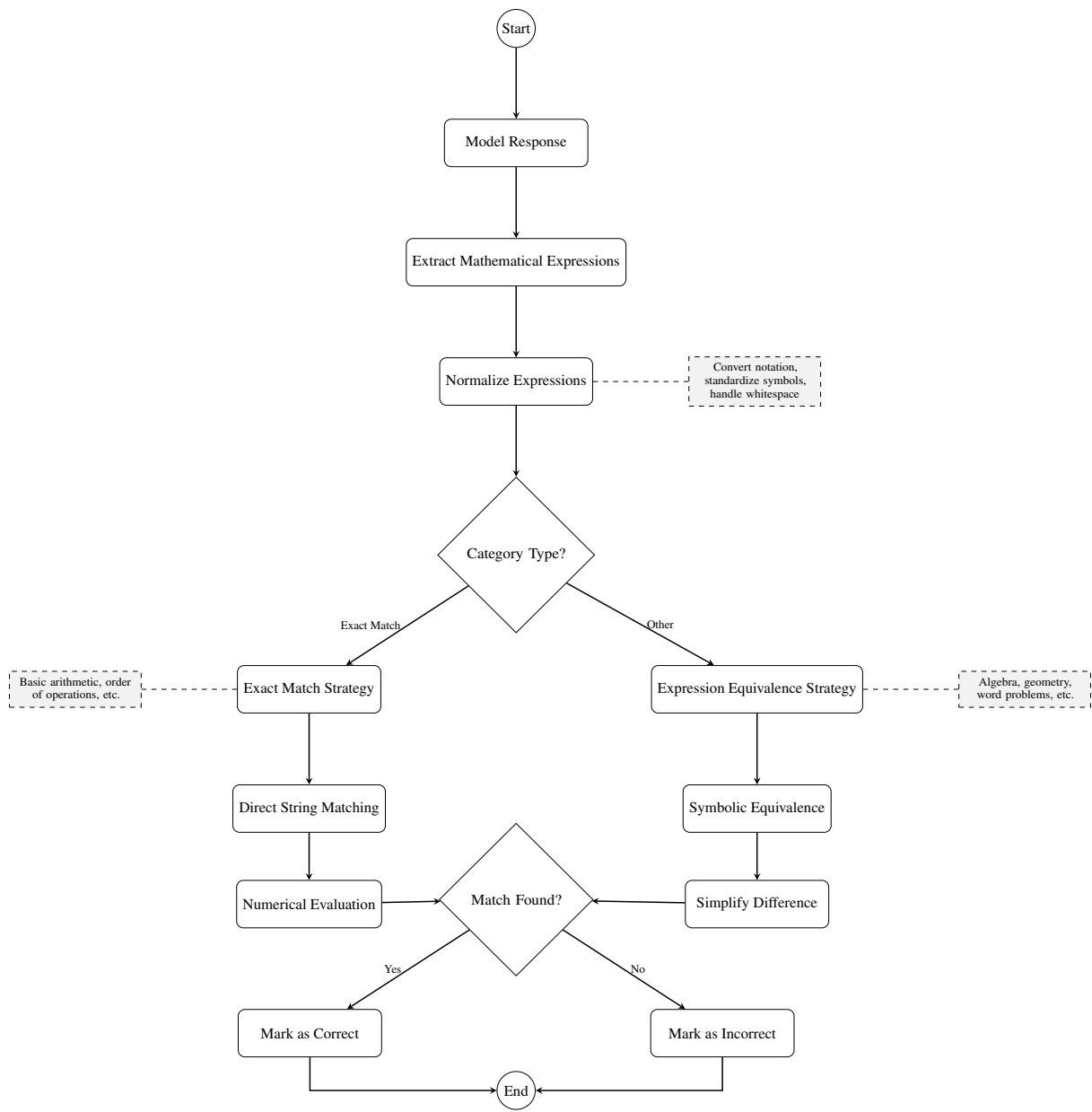
\begin{figure}[H]
    \resizebox{\textwidth}{!}{
    \centering
    \begin{tikzpicture}[
        node distance=1.5cm and 2cm,
        box/.style={rectangle, draw, rounded corners, minimum width=3cm, minimum height=1cm, text centered, font=\small},
        decision/.style={diamond, draw, text width=3cm, text centered, inner sep=0pt, font=\small},
        arrow/.style={thick, ->, >=stealth},
        note/.style={rectangle, draw, dashed, text width=2.5cm, text centered, font=\scriptsize, fill=gray!10},
        start/.style={circle, draw, minimum size=0.8cm, inner sep=0pt, font=\small},
    ]
    
    \node[start] (start) {Start};
    
    \node[box, below=of start] (model_response) {Model Response};
    \node[box, below=of model_response] (extract) {Extract Mathematical Expressions};
    \node[box, below=of extract] (normalize) {Normalize Expressions};
    \node[note, right=of normalize] (norm_note) {Convert notation, standardize symbols, handle whitespace};
    
    \node[decision, below=of normalize] (category) {Category Type?};
    \node[box, below left=of category] (exact) {Exact Match Strategy};
    \node[box, below right=of category] (equiv) {Expression Equivalence Strategy};
    \node[note, left=of exact] (exact_note) {Basic arithmetic, order of operations, etc.};
    \node[note, right=of equiv] (equiv_note) {Algebra, geometry, word problems, etc.};
    
    \node[box, below=of exact] (direct) {Direct String Matching};
    \node[box, below=1cm of direct] (numerical) {Numerical Evaluation};
    \node[box, below=of equiv] (symbolic) {Symbolic Equivalence};
    \node[box, below=1cm of symbolic] (diff) {Simplify Difference};
    
    \node[decision, below=4cm of category] (match) {Match Found?};
    \node[box, below left=of match] (correct) {Mark as Correct};
    \node[box, below right=of match] (incorrect) {Mark as Incorrect};
    
    \node[start, below=2cm of match] (end) {End};

    \draw[arrow] (start) -- (model_response);
    \draw[arrow] (model_response) -- (extract);
    \draw[arrow] (extract) -- (normalize);
    \draw[arrow] (normalize) -- (category);
    \draw[arrow] (category) -- node[left, font=\scriptsize] {Exact Match} (exact);
    \draw[arrow] (category) -- node[right, font=\scriptsize] {Other} (equiv);
    \draw[arrow] (exact) -- (direct);
    \draw[arrow] (direct) -- (numerical);
    \draw[arrow] (equiv) -- (symbolic);
    \draw[arrow] (symbolic) -- (diff);
    \draw[arrow] (numerical) -- (match);
    \draw[arrow] (diff) -- (match);
    \draw[arrow] (match) -- node[left, font=\scriptsize] {Yes} (correct);
    \draw[arrow] (match) -- node[right, font=\scriptsize] {No} (incorrect);
    \draw[arrow] (correct) |- (end);
    \draw[arrow] (incorrect) |- (end);
    
    \draw[dashed] (normalize) -- (norm_note);
    \draw[dashed] (exact) -- (exact_note);
    \draw[dashed] (equiv) -- (equiv_note);
    
    \end{tikzpicture}
    }
    \caption{Flowchart of the EasyMath benchmark evaluation process}
    \label{fig:evaluation_flowchart}
\end{figure}
\clearpage

\section{Evaluation results on the EasyMath benchmark}
We evaluated a total of 23 models (14M–4B parameters) on EasyMath. We averaged over categories rather than over the questions, which means that we assume each category has an equal weight rather than just averaging over the questions in its totality. In addition, we present a scatter plot for a subset of models, with parameter count (x-axis) against averaged accuracy (y-axis), to highlight the scaling relationship.
\begin{table}[H]
    \centering
    \textbf{Score Interpretation:} \quad
    \cellcolor{badcolor} Poor (0-24) \quad
    \cellcolor{mehcolor} Fair (25-49) \quad
    \cellcolor{goodcolor} Good (50-74) \quad
    \cellcolor{excellentcolor} Excellent (75-100)
    \resizebox{\textwidth}{!}{
        \begin{tabular}{|c|c|c|c|c|c|c|c|c|c|c|c|c|c|c|}
        \hline
        \textbf{Model} & \textbf{Average} & \textbf{B.A.} & \textbf{O.O.P.} & \textbf{F.A.D.} & \textbf{L.N.} & \textbf{E.R.} & \textbf{P} & \textbf{W.P.} & \textbf{A.E.} & \textbf{G} & \textbf{T} & \textbf{C.S.} & \textbf{E.C.} & \textbf{M.S.R.} \\
        \hline
        pythia-14m \scite{biderman2023pythiasuiteanalyzinglarge} & \scorecolor{1.85} & \scorecolor{0.0} & \scorecolor{0.0} & \scorecolor{0.0} & \scorecolor{0.0} & \scorecolor{0.0} & \scorecolor{0.0} & \scorecolor{4.0} & \scorecolor{0.0} & \scorecolor{0.0} & \scorecolor{4.0} & \scorecolor{2.0} & \scorecolor{10.0} & \scorecolor{4.0} \\
        \hline
        pythia-70m \scite{biderman2023pythiasuiteanalyzinglarge} & \scorecolor{8.31} & \scorecolor{0.0} & \scorecolor{24.0} & \scorecolor{0.0} & \scorecolor{0.0} & \scorecolor{12.0} & \scorecolor{4.0} & \scorecolor{14.0} & \scorecolor{20.0} & \scorecolor{4.0} & \scorecolor{12.0} & \scorecolor{4.0} & \scorecolor{10.0} & \scorecolor{4.0} \\
        \hline
        gpt-neo-125m \scite{gpt-neo} & \scorecolor{8.15} & \scorecolor{0.0} & \scorecolor{16.0} & \scorecolor{24.0} & \scorecolor{0.0} & \scorecolor{16.0} & \scorecolor{0.0} & \scorecolor{10.0} & \scorecolor{12.0} & \scorecolor{4.0} & \scorecolor{4.0} & \scorecolor{2.0} & \scorecolor{10.0} & \scorecolor{8.0} \\
        \hline
        SmolLM2-135M \scite{allal2025smollm2smolgoesbig} & \scorecolor{28.77} & \scorecolor{70.0} & \scorecolor{16.0} & \scorecolor{40.0} & \scorecolor{0.0} & \scorecolor{32.0} & \scorecolor{8.0} & \scorecolor{42.0} & \scorecolor{14.0} & \scorecolor{28.0} & \scorecolor{28.0} & \scorecolor{10.0} & \scorecolor{70.0} & \scorecolor{16.0} \\
        \hline
        pythia-160m \scite{biderman2023pythiasuiteanalyzinglarge} & \scorecolor{6.77} & \scorecolor{0.0} & \scorecolor{12.0} & \scorecolor{24.0} & \scorecolor{0.0} & \scorecolor{8.0} & \scorecolor{0.0} & \scorecolor{4.0} & \scorecolor{4.0} & \scorecolor{4.0} & \scorecolor{4.0} & \scorecolor{4.0} & \scorecolor{20.0} & \scorecolor{4.0} \\
        \hline
        SmolLM2-360M \scite{allal2025smollm2smolgoesbig} & \scorecolor{48.77} & \scorecolor{90.0} & \scorecolor{52.0} & \scorecolor{76.0} & \scorecolor{12.0} & \scorecolor{72.0} & \scorecolor{52.0} & \scorecolor{60.0} & \scorecolor{36.0} & \scorecolor{52.0} & \scorecolor{40.0} & \scorecolor{26.0} & \scorecolor{30.0} & \scorecolor{36.0} \\
        \hline
        pythia-410m \scite{biderman2023pythiasuiteanalyzinglarge} & \scorecolor{9.54} & \scorecolor{20.0} & \scorecolor{36.0} & \scorecolor{28.0} & \scorecolor{0.0} & \scorecolor{8.0} & \scorecolor{0.0} & \scorecolor{8.0} & \scorecolor{0.0} & \scorecolor{4.0} & \scorecolor{8.0} & \scorecolor{4.0} & \scorecolor{0.0} & \scorecolor{4.0} \\
        \hline
        Qwen2.5-0.5B \scite{qwen2025qwen25technicalreport} & \scorecolor{88.31} & \scorecolor{100.0} & \scorecolor{100.0} & \scorecolor{92.0} & \scorecolor{48.0} & \scorecolor{92.0} & \scorecolor{96.0} & \scorecolor{92.0} & \scorecolor{78.0} & \scorecolor{96.0} & \scorecolor{76.0} & \scorecolor{90.0} & \scorecolor{100.0} & \scorecolor{88.0} \\
        \hline
        Falcon3-1B \scite{Falcon3} & \scorecolor{77.69} & \scorecolor{100.0} & \scorecolor{84.0} & \scorecolor{100.0} & \scorecolor{44.0} & \scorecolor{92.0} & \scorecolor{76.0} & \scorecolor{90.0} & \scorecolor{82.0} & \scorecolor{72.0} & \scorecolor{72.0} & \scorecolor{50.0} & \scorecolor{80.0} & \scorecolor{68.0} \\
        \hline
        Llama-3.2-1B \scite{grattafiori2024llama3herdmodels,meta_llama} & \scorecolor{74.92} & \scorecolor{100.0} & \scorecolor{80.0} & \scorecolor{96.0} & \scorecolor{20.0} & \scorecolor{96.0} & \scorecolor{96.0} & \scorecolor{82.0} & \scorecolor{68.0} & \scorecolor{76.0} & \scorecolor{52.0} & \scorecolor{52.0} & \scorecolor{80.0} & \scorecolor{76.0} \\
        \hline
        Gemma-3-1B \scite{Gemma3Report2025} & \scorecolor{82.46} & \scorecolor{90.0} & \scorecolor{96.0} & \scorecolor{96.0} & \scorecolor{68.0} & \scorecolor{100.0} & \scorecolor{88.0} & \scorecolor{90.0} & \scorecolor{78.0} &   \scorecolor{84.0} & \scorecolor{44.0} & \scorecolor{72.0} & \scorecolor{90.0} & \scorecolor{76.0} \\
        \hline
        pythia-1b \scite{biderman2023pythiasuiteanalyzinglarge} & \scorecolor{11.38} & \scorecolor{20.0} & \scorecolor{20.0} & \scorecolor{28.0} & \scorecolor{12.0} & \scorecolor{16.0} & \scorecolor{8.0} & \scorecolor{6.0} & \scorecolor{10.0} & \scorecolor{4.0} & \scorecolor{8.0} & \scorecolor{6.0} & \scorecolor{10.0} & \scorecolor{0.0} \\
        \hline
        gpt-neo-1.3B \scite{gpt-neo} & \scorecolor{13.85} & \scorecolor{30.0} & \scorecolor{24.0} & \scorecolor{28.0} & \scorecolor{0.0} & \scorecolor{20.0} & \scorecolor{4.0} & \scorecolor{6.0} & \scorecolor{12.0} & \scorecolor{8.0} & \scorecolor{0.0} & \scorecolor{8.0} & \scorecolor{40.0} & \scorecolor{0.0} \\
        \hline
        pythia-1.4b \scite{biderman2023pythiasuiteanalyzinglarge} & \scorecolor{16.77} & \scorecolor{60.0} & \scorecolor{16.0} & \scorecolor{36.0} & \scorecolor{16.0} & \scorecolor{8.0} & \scorecolor{8.0} & \scorecolor{6.0} & \scorecolor{8.0} & \scorecolor{0.0} & \scorecolor{12.0} & \scorecolor{6.0} & \scorecolor{30.0} & \scorecolor{12.0} \\
        \hline
        DeepScaleR-1.5B \scite{deepscaler2025} & \scorecolor{90.00} & \scorecolor{100.0} & \scorecolor{100.0} & \scorecolor{100.0} & \scorecolor{48.0} & \scorecolor{100.0} & \scorecolor{100.0} & \scorecolor{96.0} & \scorecolor{90.0} & \scorecolor{96.0} & \scorecolor{80.0} & \scorecolor{88.0} & \scorecolor{80.0} & \scorecolor{92.0} \\
        \hline
        Qwen2.5-1.5B \scite{qwen2025qwen25technicalreport} & \scorecolor{93.08} & \scorecolor{100.0} & \scorecolor{100.0} & \scorecolor{96.0} & \scorecolor{64.0} & \scorecolor{100.0} & \scorecolor{100.0} & \scorecolor{94.0} & \scorecolor{90.0} & \scorecolor{100.0} & \scorecolor{84.0} & \scorecolor{94.0} & \scorecolor{100.0} & \scorecolor{88.0} \\
        \hline
        Qwen2.5-Math-1.5B \scite{yang2024qwen25mathtechnicalreportmathematical} & \scorecolor{95.23} & \scorecolor{100.0} & \scorecolor{100.0} & \scorecolor{100.0} & \scorecolor{72.0} & \scorecolor{100.0} & \scorecolor{100.0} & \scorecolor{100.0} & \scorecolor{98.0} & \scorecolor{100.0} & \scorecolor{88.0} & \scorecolor{94.0} & \scorecolor{90.0} & \scorecolor{96.0} \\
        \hline
        DS-R1-Distill-Qwen-1.5B \scite{deepseekai2025deepseekr1incentivizingreasoningcapability} & \scorecolor{87.85} & \scorecolor{100.0} & \scorecolor{100.0} & \scorecolor{100.0} & \scorecolor{40.0} & \scorecolor{100.0} & \scorecolor{100.0} & \scorecolor{96.0} & \scorecolor{100.0} & \scorecolor{84.0} & \scorecolor{72.0} & \scorecolor{90.0} & \scorecolor{80.0} & \scorecolor{80.0} \\
        \hline
        AceMath-1.5B \scite{liu2025acemathadvancingfrontiermath} & \scorecolor{95.69} & \scorecolor{100.0} & \scorecolor{100.0} & \scorecolor{100.0} & \scorecolor{76.0} & \scorecolor{100.0} & \scorecolor{100.0} & \scorecolor{100.0} & \scorecolor{100.0} & \scorecolor{100.0} & \scorecolor{88.0} & \scorecolor{94.0} & \scorecolor{90.0} & \scorecolor{96.0} \\
        \hline
        SmolLM2-1.7B \scite{allal2025smollm2smolgoesbig} & \scorecolor{80.62} & \scorecolor{100.0} & \scorecolor{92.0} & \scorecolor{100.0} & \scorecolor{32.0} & \scorecolor{96.0} & \scorecolor{92.0} & \scorecolor{88.0} & \scorecolor{86.0} & \scorecolor{72.0} & \scorecolor{64.0} & \scorecolor{64.0} & \scorecolor{90.0} & \scorecolor{72.0} \\
        \hline
        SmolTulu-1.7B \scite{alrashed2024smoltuluhigherlearningrate} & \scorecolor{60.46} & \scorecolor{90.0} & \scorecolor{44.0} & \scorecolor{80.0} & \scorecolor{16.0} & \scorecolor{56.0} & \scorecolor{80.0} & \scorecolor{74.0} & \scorecolor{18.0} & \scorecolor{92.0} & \scorecolor{48.0} & \scorecolor{36.0} & \scorecolor{100.0} & \scorecolor{52.0} \\
        \hline
        Llama-3.2-3B \scite{grattafiori2024llama3herdmodels,meta_llama} & \scorecolor{83.85} & \scorecolor{100.0} & \scorecolor{100.0} & \scorecolor{100.0} & \scorecolor{28.0} & \scorecolor{100.0} & \scorecolor{100.0} & \scorecolor{94.0} & \scorecolor{88.0} & \scorecolor{76.0} & \scorecolor{68.0} & \scorecolor{70.0} & \scorecolor{90.0} & \scorecolor{76.0} \\
        \hline
        Gemma-3-4B \scite{Gemma3Report2025} & \scorecolor{91.38} & \scorecolor{100.0} & \scorecolor{100.0} & \scorecolor{100.0} & \scorecolor{84.0} & \scorecolor{100.0} & \scorecolor{100.0} & \scorecolor{100.0} & \scorecolor{98.0} &      \scorecolor{88.0} & \scorecolor{68.0} & \scorecolor{72.0} & \scorecolor{90.0} & \scorecolor{88.0} \\
        \hline
        \end{tabular}
    }
    \caption{Comparison of benchmark results sorted on parameter count}
    \label{tab:benchmark_results}
\end{table}

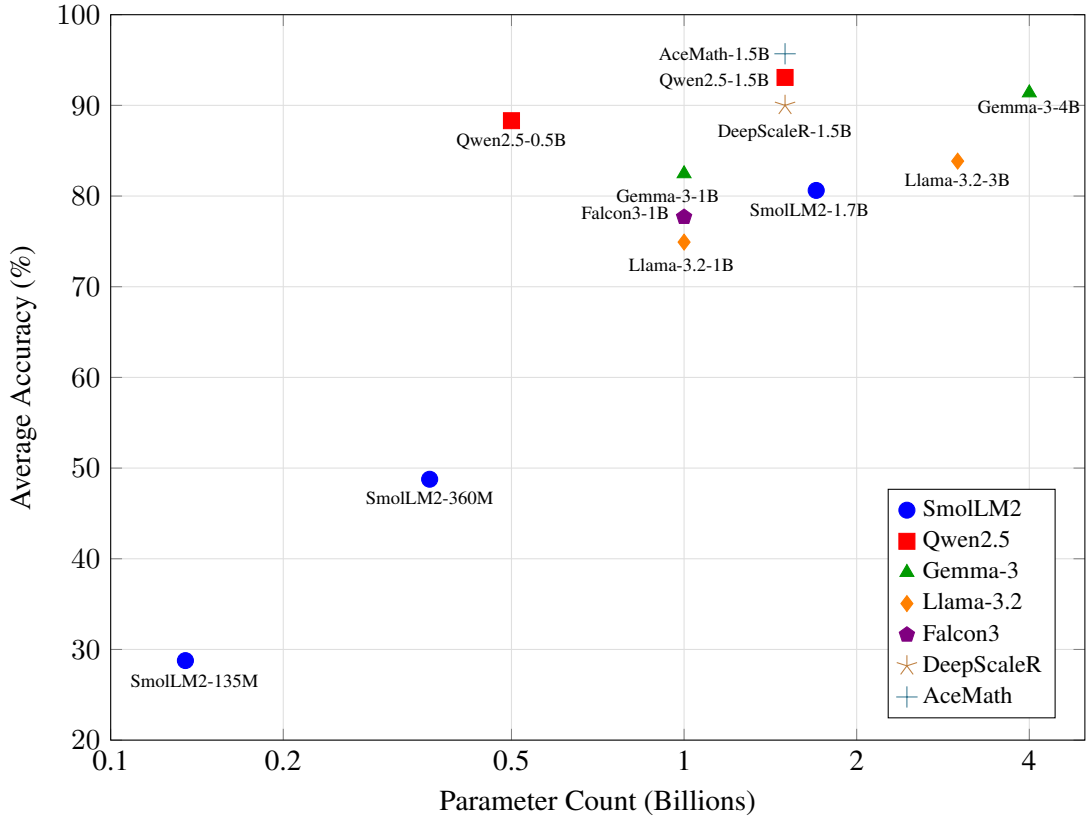
\begin{figure}[H]
    \centering
    \begin{tikzpicture}
        \begin{semilogxaxis}[
            width=0.9\textwidth,
            height=0.7\textwidth,
            xlabel={Parameter Count (Billions)},
            ylabel={Average Accuracy (\%)},
            xmin=0.1, xmax=5,
            ymin=20, ymax=100,
            grid=both,
            minor grid style={gray!25},
            major grid style={gray!25},
            legend pos=south east,
            legend style={font=\small},
            legend cell align=left,
            log ticks with fixed point,
            xtick={0.1, 0.2, 0.5, 1, 2, 4},
            xticklabels={0.1, 0.2, 0.5, 1, 2, 4},
        ]
        
        \addplot[only marks, blue, mark=*, mark size=3pt] coordinates {
            (0.135, 28.77)
            (0.36, 48.77)
            (1.7, 80.62)
        };
        \addlegendentry{SmolLM2}
        
        \addplot[only marks, red, mark=square*, mark size=3pt] coordinates {
            (0.5, 88.31)
            (1.5, 93.08)
        };
        \addlegendentry{Qwen2.5}
        
        \addplot[only marks, green!60!black, mark=triangle*, mark size=3pt] coordinates {
            (1.0, 82.46)
            (4.0, 91.38)
        };
        \addlegendentry{Gemma-3}
        
        \addplot[only marks, orange, mark=diamond*, mark size=3pt] coordinates {
            (1.0, 74.92)
            (3.0, 83.85)
        };
        \addlegendentry{Llama-3.2}
        
        \addplot[only marks, violet, mark=pentagon*, mark size=3pt] coordinates {
            (1.0, 77.69)
        };
        \addlegendentry{Falcon3}
        
        \addplot[only marks, brown, mark=star, mark size=4pt] coordinates {
            (1.5, 90.00)
        };
        \addlegendentry{DeepScaleR}
        
        \addplot[only marks, cyan!50!black, mark=+, mark size=4pt] coordinates {
            (1.5, 95.69)
        };
        \addlegendentry{AceMath}
        
        \node[anchor=south, font=\scriptsize] at (axis cs:0.14,24.7) {SmolLM2-135M};
        \node[anchor=south, font=\scriptsize] at (axis cs:0.36,44.9) {SmolLM2-360M};
        \node[anchor=south, font=\scriptsize] at (axis cs:0.5,84) {Qwen2.5-0.5B};
        \node[anchor=south, font=\scriptsize] at (axis cs:0.79,76.3) {Falcon3-1B};
        \node[anchor=south, font=\scriptsize] at (axis cs:0.99,70.6) {Llama-3.2-1B};
        \node[anchor=east, font=\scriptsize] at (axis cs:1.2,80) {Gemma-3-1B};
        \node[anchor=north, font=\scriptsize] at (axis cs:1.5,89) {DeepScaleR-1.5B};
        \node[anchor=north, font=\scriptsize] at (axis cs:1.13,94.7) {Qwen2.5-1.5B};
        \node[anchor=north, font=\scriptsize] at (axis cs:1.13,97.5) {AceMath-1.5B};
        \node[anchor=west, font=\scriptsize] at (axis cs:1.25,78.5) {SmolLM2-1.7B};
        \node[anchor=south, font=\scriptsize] at (axis cs:3,80) {Llama-3.2-3B};
        \node[anchor=south, font=\scriptsize] at (axis cs:4,88) {Gemma-3-4B};
                
    \end{semilogxaxis}
    \end{tikzpicture}
    \caption{Average accuracy vs. parameter count (log scale) across different model families}
    \label{fig:accuracy_vs_param_log}
\end{figure}
\clearpage

\section{Consistency analysis}
\begin{figure}[H]
\centering
\includegraphics[width=0.9\textwidth]{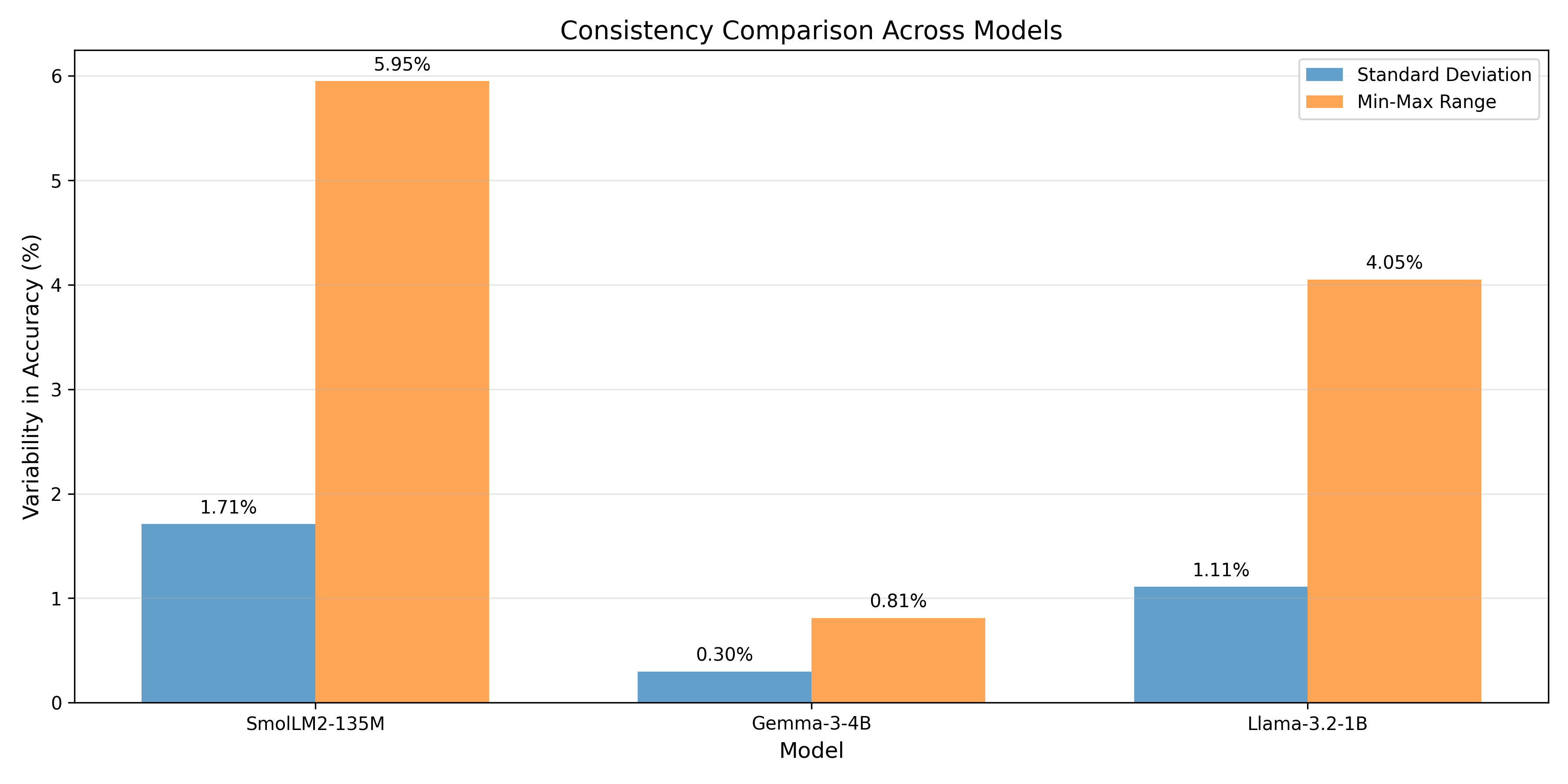}
\caption{Model consistency comparison}
\label{fig:model_consistency_comparison}
\end{figure}

\begin{figure}[H]
\centering
\includegraphics[width=0.9\textwidth]{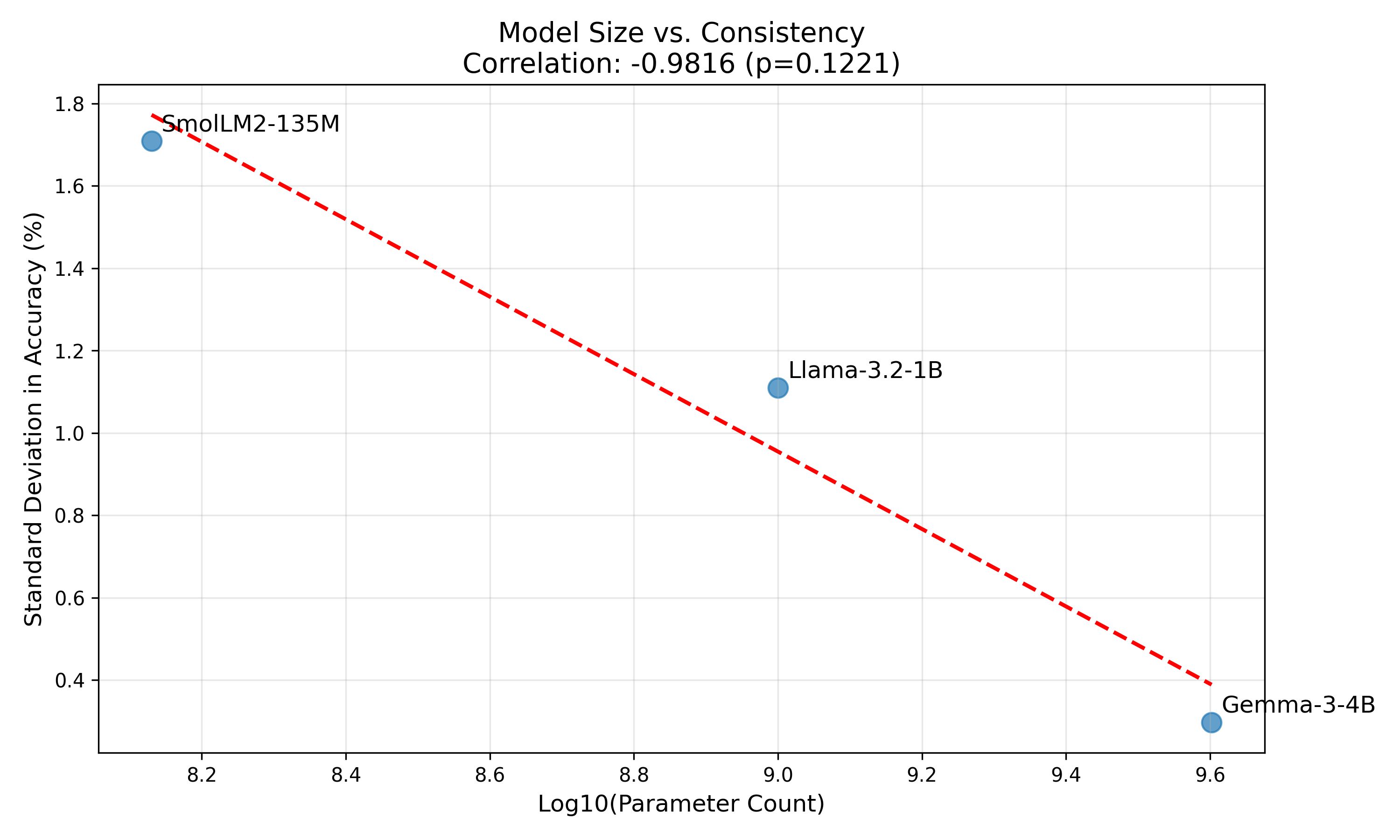}
\caption{Model size vs Consistency}
\label{fig:model_size_vs_consistency}
\end{figure}

\begin{figure}[H]
\centering
\includegraphics[width=0.9\textwidth]{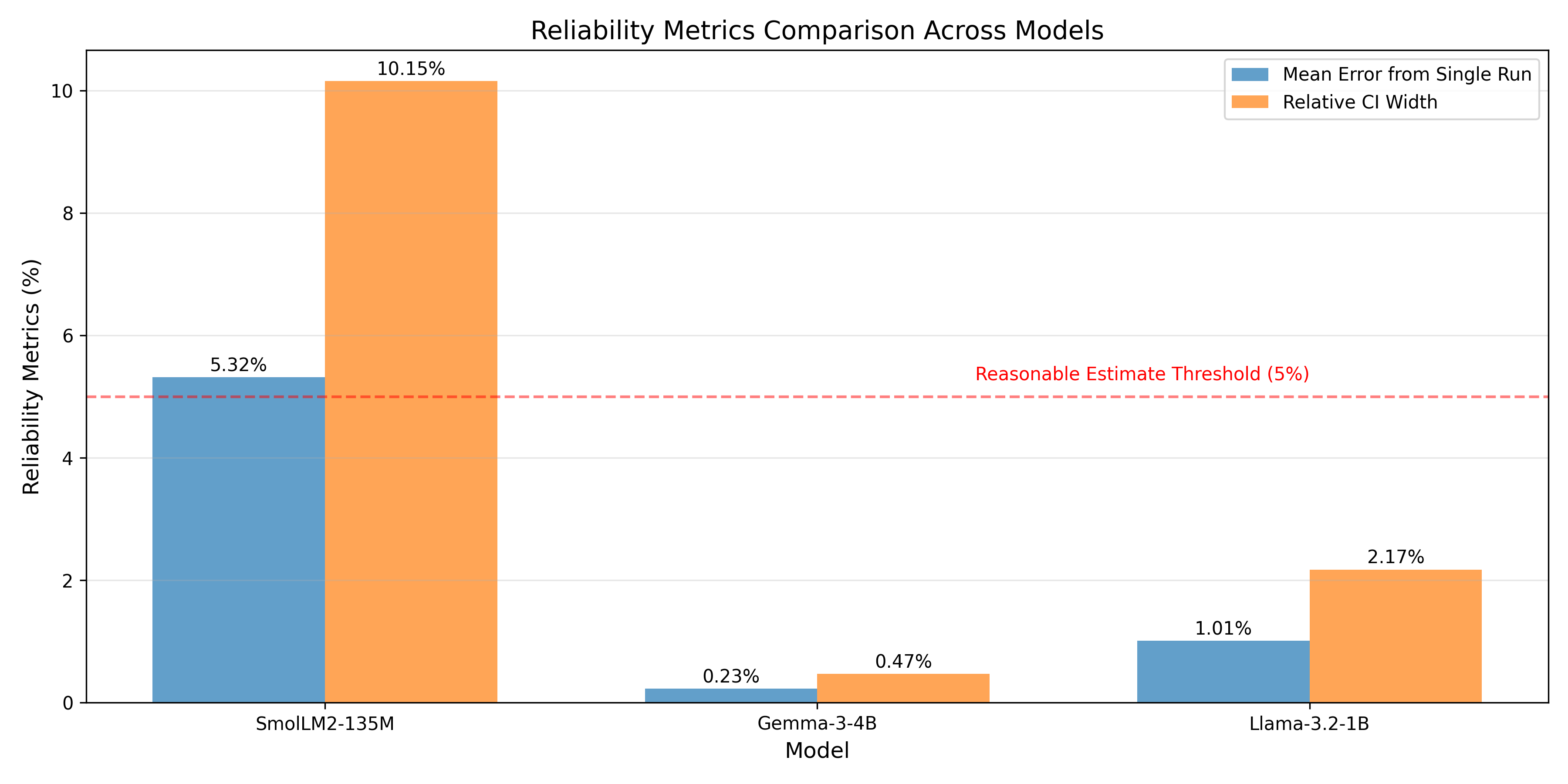}
\caption{Model reliability comparison}
\label{fig:model_reliability_comparison}
\end{figure}

\begin{figure}[H]
\centering
\includegraphics[width=0.9\textwidth]{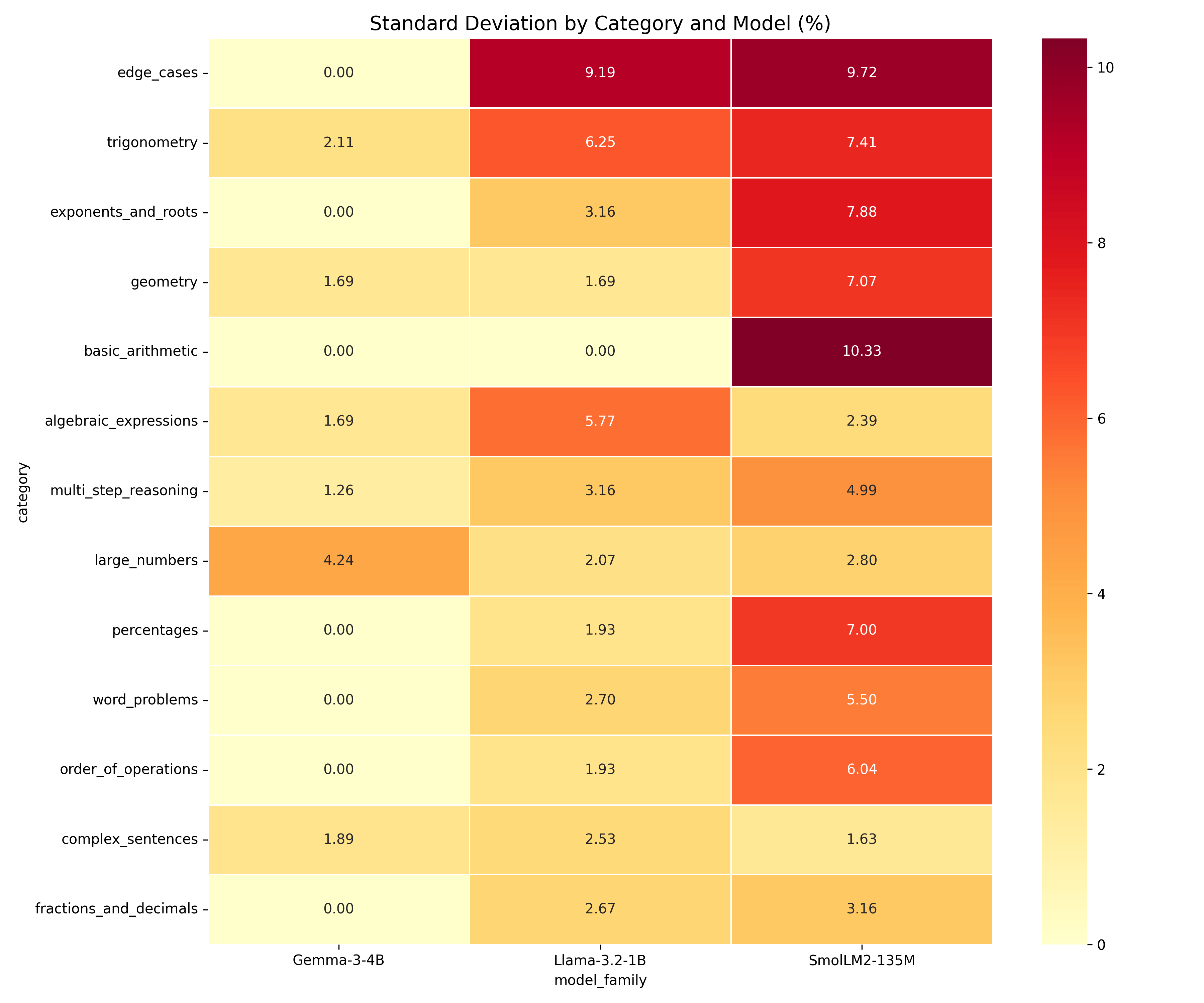}
\caption{Category variability heatmap}
\label{fig:category_variability_heatmap}
\end{figure}

We ran three models with varying parameter sizes (135M-4B) on our benchmark to assess consistency, as shown in Figures~\ref{fig:model_consistency_comparison}--\ref{fig:category_variability_heatmap}. Larger parameter models produced more consistent results, with variation increasing as parameter size decreased, supported by the strong negative correlation (-0.9816) between model size and standard deviation (Figure~\ref{fig:model_size_vs_consistency}). The heatmap (Figure~\ref{fig:category_variability_heatmap}) shows SmolLM2-135M having the highest standard deviation across categories (up to 10.33\%), while Gemma-3-4B demonstrates remarkable consistency with many 0.00\% values. As shown in Figure~\ref{fig:model_reliability_comparison}, Gemma-3-4B achieves exceptional statistical reliability (0.23\% mean error, 0.47\% CI width), while SmolLM2-135M exceeds the reasonable estimate threshold of 5\% (5.32\% mean error, 10.15\% CI width), indicating larger models require fewer evaluation runs for reliable performance assessment.
\nopagebreak

\clearpage

\clearpage

\section{Accuracy changes when training on existing datasets}
All models were fine-tuned with a learning rate of $2\!\times\!10^{-5}$, 500 warmup steps, and a cosine learning‐rate scheduler. We used the 8‐bit AdamW optimizer (``adamw\_8bit'') throughout. The data samples used were randomly sampled and not filtered in any way.
\begin{figure}[H]
  \centering
  \begin{minipage}{0.48\textwidth}
    \centering
    \begin{tikzpicture}
      \begin{axis}[
          xlabel={Epochs},
          ylabel={Average Accuracy (\%)},
          xmin=0, xmax=8,
          ymin=0, ymax=100,
          xtick={0,1,...,8},
          legend pos=north east,
          grid=both,
          legend cell align=left,
          title={SmolLM2-135M},
          width=\textwidth,
          height=0.75\textwidth
      ]
      \addplot[
        color=blue,
        mark=square*,
        thick
      ]
      coordinates {
        (0,24.05) (1,33.78) (2,36.49) (3,43.51) (4,42.16) (5,42.43) (6,40.81) (7,40.00) (8,42.16)
      };
      \addlegendentry{Orca-Math-Problems}
      
      \addplot[
        color=red,
        mark=triangle*,
        thick
      ]
      coordinates {
        (0,24.05) (1,24.05) (2,28.92) (3,29.46) (4,30.0) (5,30.0) (6,30.0) (7,28.65) (8,33.24)
      };
      \addlegendentry{R1-Math}
      
      \end{axis}
    \end{tikzpicture}
  \end{minipage}
  \hfill
  \begin{minipage}{0.48\textwidth}
    \centering
    \begin{tikzpicture}
      \begin{axis}[
          xlabel={Epochs},
          ylabel={Average Accuracy (\%)},
          xmin=0, xmax=8,
          ymin=0, ymax=100,
          xtick={0,1,...,8},
          legend pos=north east,
          grid=both,
          legend cell align=left,
          title={SmolLM2-360M},
          width=\textwidth,
          height=0.75\textwidth
      ]
      \addplot[
        color=blue,
        mark=square*,
        thick
      ]
      coordinates {
        (0,46.22) (1,58.65) (2,59.73) (3,62.97) (4,60.81) (5,60.54) (6,62.70) (7,60.27) (8,61.08)
      };
      \addlegendentry{Orca-Math-Problems}
      
      \addplot[
        color=red,
        mark=triangle*,
        thick
      ]
      coordinates {
        (0,46.22) (1,43.51) (2,52.97) (3,47.57) (4,53.51) (5,50.81) (6,47.57) (7,47.84) (8,46.49)
      };
      \addlegendentry{R1-Math}
      
      \end{axis}
    \end{tikzpicture}
  \end{minipage}
  \caption{Average accuracy (\%) over epochs for Orca-Math-Problems and R1-Math on different base-models.}
  \label{fig:accuracy_plot_finetunes}
\end{figure}
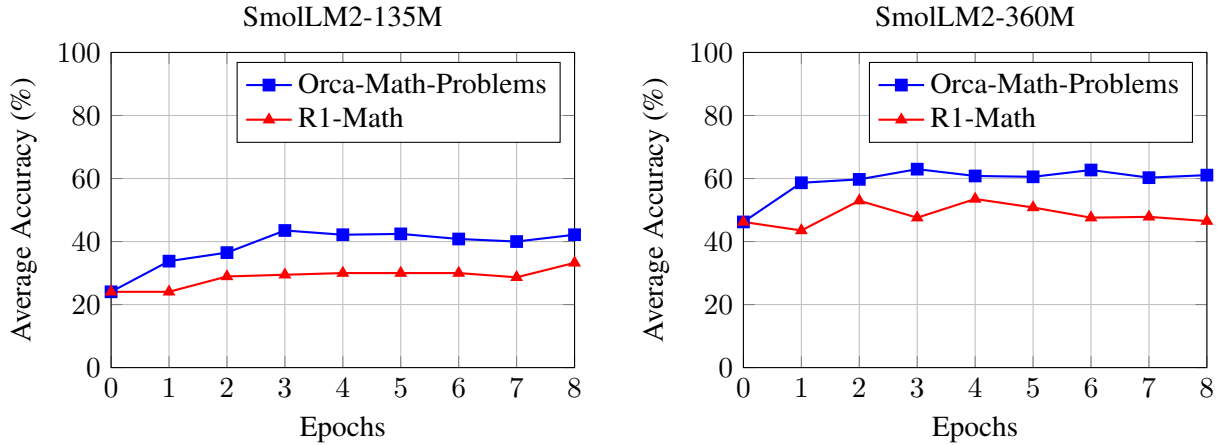

\vspace{-1em}
\begin{table}[H]
    \centering
    \textbf{Score Interpretation:} \quad
    \cellcolor{badcolor} Poor (0-24) \quad
    \cellcolor{mehcolor} Fair (25-49) \quad
    \cellcolor{goodcolor} Good (50-74) \quad
    \cellcolor{excellentcolor} Excellent (75-100)
    \resizebox{\textwidth}{!}{
        \begin{tabular}{|c|c|c|c|c|c|c|c|c|c|c|c|c|c|c|}
        \hline
        \textbf{Epoch} & \textbf{Average} & \textbf{B.A.} & \textbf{O.O.P.} & \textbf{F.A.D.} & \textbf{L.N.} & \textbf{E.R.} & \textbf{P} & \textbf{W.P.} & \textbf{A.E.} & \textbf{G} & \textbf{T} & \textbf{C.S.} & \textbf{E.C.} & \textbf{M.S.R.} \\
        \hline
        Baseline & \scorecolor{24.05} & \scorecolor{70.0} & \scorecolor{16.0} & \scorecolor{40.0} & \scorecolor{0.0} & \scorecolor{32.0} & \scorecolor{8.0} & \scorecolor{42.0} & \scorecolor{14.0} & \scorecolor{28.0} & \scorecolor{28.0} & \scorecolor{10.0} & \scorecolor{70.0} & \scorecolor{16.0} \\
        \hline
        Epoch 2 & \scorecolor{36.49} & \scorecolor{80.0} & \scorecolor{32.0} & \scorecolor{76.0} & \scorecolor{0.0} & \scorecolor{56.0} & \scorecolor{68.0} & \scorecolor{58.0} & \scorecolor{14.0} & \scorecolor{36.0} & \scorecolor{24.0} & \scorecolor{24.0} & \scorecolor{0.0} & \scorecolor{24.0} \\
        \hline
        Epoch 4 & \scorecolor{42.16} & \scorecolor{70.0} & \scorecolor{40.0} & \scorecolor{76.0} & \scorecolor{4.0} & \scorecolor{64.0} & \scorecolor{76.0} & \scorecolor{70.0} & \scorecolor{20.0} & \scorecolor{44.0} & \scorecolor{20.0} & \scorecolor{28.0} & \scorecolor{0.0} & \scorecolor{36.0} \\
        \hline
        Epoch 6 & \scorecolor{40.81} & \scorecolor{90.0} & \scorecolor{44.0} & \scorecolor{80.0} & \scorecolor{0.0} & \scorecolor{64.0} & \scorecolor{60.0} & \scorecolor{72.0} & \scorecolor{20.0} & \scorecolor{44.0} & \scorecolor{24.0} & \scorecolor{24.0} & \scorecolor{10.0} & \scorecolor{16.0} \\
        \hline
        Epoch 8 & \scorecolor{42.16} & \scorecolor{60.0} & \scorecolor{52.0} & \scorecolor{68.0} & \scorecolor{4.0} & \scorecolor{56.0} & \scorecolor{68.0} & \scorecolor{62.0} & \scorecolor{20.0} & \scorecolor{44.0} & \scorecolor{36.0} & \scorecolor{34.0} & \scorecolor{20.0} & \scorecolor{32.0} \\
        \hline
        \end{tabular}
    }
    \caption{SmolLM2-135M-Instruct finetuned on Orca-Math-Problems. Averaged over all questions.}
    \label{tab:smol_135m_orca}
\end{table}

\vspace{-1em}
\begin{table}[H]
    \centering
    \textbf{Score Interpretation:} \quad
    \cellcolor{badcolor} Poor (0-24) \quad
    \cellcolor{mehcolor} Fair (25-49) \quad
    \cellcolor{goodcolor} Good (50-74) \quad
    \cellcolor{excellentcolor} Excellent (75-100)
    \resizebox{\textwidth}{!}{
        \begin{tabular}{|c|c|c|c|c|c|c|c|c|c|c|c|c|c|c|}
        \hline
        \textbf{Epoch} & \textbf{Average} & \textbf{B.A.} & \textbf{O.O.P.} & \textbf{F.A.D.} & \textbf{L.N.} & \textbf{E.R.} & \textbf{P} & \textbf{W.P.} & \textbf{A.E.} & \textbf{G} & \textbf{T} & \textbf{C.S.} & \textbf{E.C.} & \textbf{M.S.R.} \\
        \hline
        Baseline & \scorecolor{46.22} & \scorecolor{90.0} & \scorecolor{52.0} & \scorecolor{76.0} & \scorecolor{12.0} & \scorecolor{72.0} & \scorecolor{52.0} & \scorecolor{60.0} & \scorecolor{36.0} & \scorecolor{52.0} & \scorecolor{40.0} & \scorecolor{26.0} & \scorecolor{30.0} & \scorecolor{36.0} \\
        \hline
        Epoch 2 & \scorecolor{59.73} & \scorecolor{100.0} & \scorecolor{80.0} & \scorecolor{80.0} & \scorecolor{16.0} & \scorecolor{76.0} & \scorecolor{84.0} & \scorecolor{82.0} & \scorecolor{38.0} & \scorecolor{64.0} & \scorecolor{40.0} & \scorecolor{42.0} & \scorecolor{70.0} & \scorecolor{52.0} \\
        \hline
        Epoch 4 & \scorecolor{60.81} & \scorecolor{100.0} & \scorecolor{72.0} & \scorecolor{84.0} & \scorecolor{20.0} & \scorecolor{80.0} & \scorecolor{84.0} & \scorecolor{88.0} & \scorecolor{30.0} & \scorecolor{64.0} & \scorecolor{32.0} & \scorecolor{50.0} & \scorecolor{80.0} & \scorecolor{56.0} \\
        \hline
        Epoch 6 & \scorecolor{62.70} & \scorecolor{100.0} & \scorecolor{76.0} & \scorecolor{84.0} & \scorecolor{28.0} & \scorecolor{84.0} & \scorecolor{96.0} & \scorecolor{82.0} & \scorecolor{28.0} & \scorecolor{68.0} & \scorecolor{48.0} & \scorecolor{48.0} & \scorecolor{90.0} & \scorecolor{52.0} \\
        \hline
        Epoch 8 & \scorecolor{61.08} & \scorecolor{100.0} & \scorecolor{76.0} & \scorecolor{80.0} & \scorecolor{20.0} & \scorecolor{80.0} & \scorecolor{96.0} & \scorecolor{76.0} & \scorecolor{34.0} & \scorecolor{52.0} & \scorecolor{48.0} & \scorecolor{54.0} & \scorecolor{80.0} & \scorecolor{52.0} \\
        \hline
        \end{tabular}
    }
    \caption{SmolLM2-360M-Instruct finetuned on Orca-Math-Problems. Averaged over all questions.}
    \label{tab:smol_360m_orca}
\end{table}

\vspace{-1em}
\begin{table}[H]
    \centering
    \textbf{Score Interpretation:} \quad
    \cellcolor{badcolor} Poor (0-24) \quad
    \cellcolor{mehcolor} Fair (25-49) \quad
    \cellcolor{goodcolor} Good (50-74) \quad
    \cellcolor{excellentcolor} Excellent (75-100)
    \resizebox{\textwidth}{!}{
        \begin{tabular}{|c|c|c|c|c|c|c|c|c|c|c|c|c|c|c|}
        \hline
        \textbf{Epoch} & \textbf{Average} & \textbf{B.A.} & \textbf{O.O.P.} & \textbf{F.A.D.} & \textbf{L.N.} & \textbf{E.R.} & \textbf{P} & \textbf{W.P.} & \textbf{A.E.} & \textbf{G} & \textbf{T} & \textbf{C.S.} & \textbf{E.C.} & \textbf{M.S.R.} \\
        \hline
        Baseline & \scorecolor{24.05} & \scorecolor{70.0} & \scorecolor{16.0} & \scorecolor{40.0} & \scorecolor{0.0} & \scorecolor{32.0} & \scorecolor{8.0} & \scorecolor{42.0} & \scorecolor{14.0} & \scorecolor{28.0} & \scorecolor{28.0} & \scorecolor{10.0} & \scorecolor{70.0} & \scorecolor{16.0} \\
        \hline
        Epoch 2 & \scorecolor{28.92} & \scorecolor{70.0} & \scorecolor{36.0} & \scorecolor{44.0} & \scorecolor{8.0} & \scorecolor{48.0} & \scorecolor{12.0} & \scorecolor{24.0} & \scorecolor{18.0} & \scorecolor{36.0} & \scorecolor{32.0} & \scorecolor{28.0} & \scorecolor{70.0} & \scorecolor{16.0} \\
        \hline
        Epoch 4 & \scorecolor{30.00} & \scorecolor{80.0} & \scorecolor{48.0} & \scorecolor{40.0} & \scorecolor{4.0} & \scorecolor{64.0} & \scorecolor{8.0} & \scorecolor{36.0} & \scorecolor{22.0} & \scorecolor{28.0} & \scorecolor{20.0} & \scorecolor{30.0} & \scorecolor{30.0} & \scorecolor{12.0} \\
        \hline
        Epoch 6 & \scorecolor{30.0} & \scorecolor{40.0} & \scorecolor{52.0} & \scorecolor{52.0} & \scorecolor{0.0} & \scorecolor{60.0} & \scorecolor{20.0} & \scorecolor{32.0} & \scorecolor{20.0} & \scorecolor{24.0} & \scorecolor{24.0} & \scorecolor{28.0} & \scorecolor{50.0} & \scorecolor{16.0} \\
        \hline
        Epoch 8 & \scorecolor{33.24} & \scorecolor{70.0} & \scorecolor{60.0} & \scorecolor{36.0} & \scorecolor{0.0} & \scorecolor{52.0} & \scorecolor{24.0} & \scorecolor{38.0} & \scorecolor{28.0} & \scorecolor{32.0} & \scorecolor{44.0} & \scorecolor{28.0} & \scorecolor{30.0} & \scorecolor{16.0} \\
        \hline
        \end{tabular}
    }
    \caption{SmolLM2-135M-Instruct finetuned on R1-Math. Averaged over all questions.}
    \label{tab:smol_135m_r1}
\end{table}

\vspace{-1em}
\begin{table}[H]
    \centering
    \textbf{Score Interpretation:} \quad
    \cellcolor{badcolor} Poor (0-24) \quad
    \cellcolor{mehcolor} Fair (25-49) \quad
    \cellcolor{goodcolor} Good (50-74) \quad
    \cellcolor{excellentcolor} Excellent (75-100)
    \resizebox{\textwidth}{!}{
        \begin{tabular}{|c|c|c|c|c|c|c|c|c|c|c|c|c|c|c|}
        \hline
        \textbf{Epoch} & \textbf{Average} & \textbf{B.A.} & \textbf{O.O.P.} & \textbf{F.A.D.} & \textbf{L.N.} & \textbf{E.R.} & \textbf{P} & \textbf{W.P.} & \textbf{A.E.} & \textbf{G} & \textbf{T} & \textbf{C.S.} & \textbf{E.C.} & \textbf{M.S.R.} \\
        \hline
        Baseline & \scorecolor{46.22} & \scorecolor{90.0} & \scorecolor{52.0} & \scorecolor{76.0} & \scorecolor{12.0} & \scorecolor{72.0} & \scorecolor{52.0} & \scorecolor{60.0} & \scorecolor{36.0} & \scorecolor{52.0} & \scorecolor{40.0} & \scorecolor{26.0} & \scorecolor{30.0} & \scorecolor{36.0} \\
        \hline
        Epoch 2 & \scorecolor{52.97} & \scorecolor{100.0} & \scorecolor{76.0} & \scorecolor{76.0} & \scorecolor{8.0} & \scorecolor{88.0} & \scorecolor{52.0} & \scorecolor{64.0} & \scorecolor{38.0} & \scorecolor{36.0} & \scorecolor{52.0} & \scorecolor{52.0} & \scorecolor{50.0} & \scorecolor{28.0} \\
        \hline
        Epoch 4 & \scorecolor{53.51} & \scorecolor{100.0} & \scorecolor{80.0} & \scorecolor{52.0} & \scorecolor{20.0} & \scorecolor{92.0} & \scorecolor{56.0} & \scorecolor{64.0} & \scorecolor{34.0} & \scorecolor{52.0} & \scorecolor{60.0} & \scorecolor{34.0} & \scorecolor{90.0} & \scorecolor{40.0} \\
        \hline
        Epoch 6 & \scorecolor{47.57} & \scorecolor{100.0} & \scorecolor{72.0} & \scorecolor{72.0} & \scorecolor{16.0} & \scorecolor{84.0} & \scorecolor{52.0} & \scorecolor{56.0} & \scorecolor{34.0} & \scorecolor{40.0} & \scorecolor{36.0} & \scorecolor{26.0} & \scorecolor{60.0} & \scorecolor{36.0} \\
        \hline
        Epoch 8 & \scorecolor{46.49} & \scorecolor{90.0} & \scorecolor{64.0} & \scorecolor{72.0} & \scorecolor{8.0} & \scorecolor{84.0} & \scorecolor{36.0} & \scorecolor{60.0} & \scorecolor{32.0} & \scorecolor{16.0} & \scorecolor{44.0} & \scorecolor{38.0} & \scorecolor{50.0} & \scorecolor{48.0} \\
        \hline
        \end{tabular}
    }
    \caption{SmolLM2-360M-Instruct finetuned on R1-Math. Averaged over all questions.}
    \label{tab:smol_360m_r1}
\end{table}
\clearpage

\section{Prompt engineering example - pythia-1.4b}
To discover the effect prompting can have on the achieved accuracy we prompted pythia-1.4b \cite{biderman2023pythiasuiteanalyzinglarge} using a standard QA format and separately with a CoT prompt.
\nopagebreak
\begin{table}[H]
  \centering
  \begin{tabular}{lll}
    \hline
    \textbf{Categories}           & \textbf{Standard QA format} & \textbf{CoT prompt} \\
    \hline
    basic-arithmetic              & \textbf{60.0}               & 50.0                \\
    order-of-operations           & 16.0                        & \textbf{24.0}       \\
    fractions-and-decimals        & 36.0                        & 36.0                \\
    large-numbers                 & 16.0                        & \textbf{20.0}       \\
    exponents-and-roots           & 8.0                         & \textbf{24.0}       \\
    percentages                   & 8.0                         & \textbf{28.0}       \\
    word-problems                 & 6.0                         & \textbf{28.0}       \\
    algebraic-expressions         & 8.0                         & \textbf{12.0}       \\
    geometry                      & 0.0                         & \textbf{4.0}        \\
    trigonometry                  & 12.0                        & 12.0                \\
    complex-sentences             & 6.0                         & \textbf{24.0}       \\
    edge-cases                    & \textbf{30.0}               & 20.0                \\
    multi-step-reasoning          & 12.0                        & \textbf{16.0}       \\
    \hline
    \textbf{Average}              & 16.77                       & \textbf{22.92}      \\
    \hline
  \end{tabular}
  \caption{
    Prompt engineering comparison. We average over the categories rather than the question due to varying amounts of questions.
  }
  \label{tab:pythia-easymath-scores}
\end{table}

\subsection*{Prompt Templates}

The prompt templates that were used:

First, the Q–A prompt:
\begin{tcolorbox}[enhanced,
                  colback=gray!10,
                  colframe=gray!50,
                  fonttitle=\bfseries,
                  title=Q–A Prompt Template]
Question: \{question\} \\
Answer:
\end{tcolorbox}

Next, the Chain-of-Thought prompt:
\begin{tcolorbox}[enhanced,
                  colback=gray!10,
                  colframe=gray!50,
                  fonttitle=\bfseries,
                  title=CoT Prompt Template]
Question: \{question\} \\
I will now reason step by step to solve this. Answer:
\end{tcolorbox}

\section{Benchmark comparison}
We present the following table, showing results from testing six models on five benchmarks, to evaluate how well each benchmark differentiates model performance.
\nopagebreak
\begin{table}[H]
  \centering
  \begin{tabular}{llllll}
    \hline
    \textbf{Models} & \textbf{GSM8K} & \textbf{hendrycks-MATH} & \textbf{MATH-500} & \textbf{MathQA} & \textbf{EasyMath} \\    
    \hline
    pythia-70m & 00.0 & 00.0 & 00.0 & 18.9 & 8.31 \\
    SmolLM2-135M & 00.0 & 00.0 & 00.0 & 21.3 & 28.77 \\
    Qwen2.5-0.5B & 00.0 & 11.3 & 13.0 & 26.7 & 88.31 \\
    Falcon3-1B & 00.0 & 7.6 & 7.6 & 26.9 & 77.69 \\
    AceMath-1.5B & 00.0 & 32.0 & 34.0 & 32.2 & 95.69 \\
    Llama-3.2-3B & 00.0 & 35.2 & 36.2 & 23.8 & 83.85 \\
    \hline
  \end{tabular}
  \caption{Benchmark 0-shot accuracy scores}
\end{table}

\end{document}